\documentclass[10pt,twocolumn,letterpaper]{article}

\usepackage{cvpr}              

\usepackage{tcolorbox}
\usepackage{listings}  
\usepackage{xcolor}    
\tcbuselibrary{listingsutf8,breakable}
\usepackage{subcaption}
\usepackage{multicol}

\definecolor{cvprblue}{rgb}{0.21,0.49,0.74}
\usepackage[pagebackref,breaklinks,colorlinks,allcolors=cvprblue]{hyperref}
\usepackage[accsupp]{axessibility}  

\usepackage{url}

\usepackage{algorithm}
\usepackage{algorithmic}
\usepackage{booktabs}
\usepackage{times}
\usepackage{epsfig}
\usepackage{amsmath}
\usepackage{amssymb}

\usepackage{multirow}
\usepackage[flushleft]{threeparttable}
\usepackage{pifont}

\usepackage{color}
\usepackage{array}
\usepackage{makecell}

\newlength\savewidth
\usepackage[utf8]{inputenc} 
\usepackage[T1]{fontenc}    
\usepackage{hyperref}       
\usepackage{amsfonts}       
\usepackage{nicefrac}       
\usepackage{microtype}      
\usepackage{xcolor}         

\usepackage{tabularray}
\usepackage{colortbl}
\usepackage{arydshln}

\definecolor{mygray}{gray}{.9}
\definecolor{mygreen}{rgb}{0, 0.6, 0}
\definecolor{myteal}{rgb}{0.16, 0.47, 0.56}
\definecolor{mypink}{rgb}{0.81, 0.25, 0.44}

\newcommand{\improve}[1]{\scriptsize{\textcolor[RGB]{61,145,64}{~(#1)}}}

\def\paperID{6627} 
\def\confName{CVPR}
\def\confYear{2025}

\title{
STPro: Spatial and Temporal Progressive Learning for Weakly Supervised Spatio-Temporal Grounding\\
}

\author{Aaryan Garg$^{1}$ \quad Akash Kumar$^{2}$ \quad Yogesh S Rawat$^{2}$  \\
$^{1}$BITS Pilani \quad $^{2}$CRCV, University of Central Florida\\
{\tt\small f20212222@pilani.bits-pilani.ac.in} \quad {\tt\small \{akash.kumar, yogesh\}@ucf.edu} \\
\normalsize Project Page: \url{https://aaryangrg.github.io/research/stpro}  \\
}

\begin{document}

\maketitle

\begin{abstract}
In this work we study Weakly Supervised Spatio-Temporal Video Grounding (WSTVG), a challenging task of localizing subjects spatio-temporally in videos using only textual queries and no bounding box supervision. Inspired by recent advances in vision-language foundation models, we investigate their utility for WSTVG, leveraging their zero-shot grounding capabilities. However, we find that a simple adaptation lacks essential spatio-temporal grounding abilities. To bridge this gap, we introduce \textbf{Tubelet Referral Grounding (TRG)}, which connects textual queries to tubelets to enable spatio-temporal predictions. Despite its promise, TRG struggles with compositional action understanding and dense scene scenarios. To address these limitations, we propose \textbf{STPro}, a novel progressive learning framework with two key modules: (1) \textit{Sub-Action Temporal Curriculum Learning (SA-TCL)}, which incrementally builds compositional action understanding, and (2) \textit{Congestion-Guided Spatial Curriculum Learning (CG-SCL)}, which adapts the model to complex scenes by spatially increasing task difficulty. STPro achieves state-of-the-art results on three benchmark datasets, with improvements of 1.0\% on VidSTG-Declarative and 3.0\% on HCSTVG-v1. 

\end{abstract}
    
\section{Introduction}
\label{sec:intro}
Spatio-temporal video grounding (STVG) seeks to detect and localize objects within video sequences across both space and time, guided solely by textual descriptions. This capability is crucial for applications like surveillance, autonomous navigation, and scene understanding. However, STVG presents significant challenges, requiring models to distinguish target objects from distractors over time and to identify precise temporal boundaries of actions. While recent methods tackle STVG in fully-supervised settings \citep{Yang2022TubeDETRSV, Jin2022EmbracingCA, video-gdino}, they rely on costly, labor-intensive frame-level annotations. In this work, we address these challenges through weakly supervised STVG (WSTVG), training models with high-level video descriptions and eliminating the need for detailed spatio-temporal labels.

Existing research in weakly supervised learning for grounding has focused primarily on image-based phrase \cite{datta2019align2ground, wang-etal-2020-maf, liu2021relation} and referral grounding \cite{liu2019adaptive, liu2022entity, refclip}. More recently, studies have extended weak supervision to the STVG task \cite{nafae, Chen2019WeaklySupervisedSG, Li_2023_CVPR, vcma}. However, these approaches are often complex, involving multiple hierarchical algorithms \citep{Li_2023_CVPR}, additional modality information \citep{Chen2019WeaklySupervisedSG}, and restricted object categories \citep{nafae, Chen2019WeaklySupervisedSG}. To address these limitations, we propose a streamlined approach that relies solely on the vision modality and supports \textit{free-form} text queries.

Recent advancements in vision-language foundation models (VFMs) \cite{madan2024foundation} have shown remarkable zero-shot grounding abilities in image-based referral tasks, driven by large-scale pretraining on diverse multimodal datasets \cite{uninext-lvlm, llavag-mllm, Liu2023GroundingDM}. These VFMs excel at associating visual entities with complex, free-form textual descriptions, making them highly effective in scenarios where image-based object localization must generalize across a variety of contexts. Grounding DINO (G-DINO) \cite{Liu2023GroundingDM}, one of the leading VFMs, has demonstrated robust zero-shot performance on numerous image-level grounding benchmarks. However, while recent work has adapted G-DINO for fully supervised video grounding \cite{video-gdino}, the potential of VFMs remains unexplored in weakly supervised spatio-temporal video grounding (WSTVG), where precise annotations are not available.

Applying image-based VFMs directly to videos introduces significant challenges. Unlike static images, videos often contain target subjects that are only visible in certain frames, with dynamic interactions that require an understanding of both spatial and temporal continuity. Simply extending G-DINO to videos results in noisy training due to irrelevant frames and complex action sequences that it cannot accurately localize. To address these limitations, we introduce the \textbf{Tubelet Referral Grounding (TRG)} module, which adapts image-based VFMs to spatio-temporal contexts by linking textual queries with tubelets (spatio-temporal object proposals). While TRG provides a strong baseline for adapting VFMs to WSTVG, it has two main shortcomings: it struggles with complex free-form queries involving actor-action relationships and lacks an ability to handle dense scenes with ordered action compositions, limiting its effectiveness in complex scenarios.

To overcome these challenges, we propose \textbf{Spatial and Temporal Progressive Learning (STPro)}, a framework specifically designed to enhance VFMs for WSTVG tasks. STPro introduces two progressive learning strategies: (1) \textit{Sub-Action Temporal Curriculum Learning (SA-TCL)}, which incrementally develops the model’s understanding of complex action sequences by gradually increasing task difficulty over time, and (2) \textit{Congestion-Guided Spatial Curriculum Learning (CG-SCL)}, which focuses the model on progressively sparse scenes. Together, these strategies help the model adapt to intricate actor-action relationships and noisy, free-form textual descriptions, significantly enhancing the VFM’s capability to handle spatio-temporal grounding in weakly supervised settings.

In summary, we make the following contributions:
\begin{itemize}
    \item We introduce \textit{Temporal Referral Grounding (TRG)} module, which adapts foundation models for the weakly supervised spatio-temporal video grounding task, providing a strong baseline for handling temporal and spatial contexts.
    \item We propose the \textit{Sub-Action Temporal Curriculum Learning (SA-TCL)} paradigm, which improves the model's temporal grounding capability by incrementally increasing the complexity of action sequences, enabling better handling of temporal dependencies in videos.
    \item We introduce \textit{Congestion-Guided Spatial Curriculum Learning (CG-SCL)}, a novel approach that progressively adapts the model to more complex and dense scene structures, improving its ability to understand spatial relationships in challenging video scenarios.
\end{itemize}
We perform our experiments on three different benchmark datasets, ViDSTG and HCSTVG-v1 and HCSTVG-v2 demonstrating effectiveness of our proposed approach. STPro outperform previous state-of-the-art methods on WSTVG task by an absolute margin of 1.0\% on VidSTG-Declarative and 3.0\% on HCSTVG-v1.
\section{Related Work}

\noindent{\textbf{Spatio-Temporal Video Grounding:}} 
It requires grounding video tubelets (detections across time) to textual queries, aligning both spatial and temporal dimensions. Traditional methods use a two-stage pipeline, separating spatial \citep{Rohrbach2015GroundingOT, stpr} and temporal grounding \citep{Gao2017TALLTA} but rely on fixed object categories, limiting adaptability for free-form queries. Recent multimodal approaches \citep{stvgbert, Yang2022TubeDETRSV, Jin2022EmbracingCA} combine image detectors, video encoders, and spatio-temporal decoders to address feature alignment and leverage both static and motion cues \citep{clb, cg-stvg}. These methods typically require extensive frame-level annotations. \textbf{\textit{In contrast}}, we remove the need for spatio-temporal labels, greatly reducing annotation cost.

\noindent{\textbf{Weakly Supervised Learning}:} Existing works on dense tasks \citep{kumar2024stable, Singh_Rana_Kumar_Vyas_Rawat_2024, kumar2023benchmarking, Rana_2023_CVPR, Kumar_2022_CVPR, ayushneurips22, Dave_2022_WACV, modi2022video} are unimodal, rely on semi-supervised or active learning, and cannot be extended to WSSTVG. Grounding techniques fall into three categories: phrase grounding \citep{rohrbach2016grounding, datta2019align2ground} (identifying objects from text using margin-based losses \citep{datta2019align2ground}, contrastive learning \citep{gupta2020contrastive}, and reconstruction \citep{rohrbach2016grounding}); referral grounding \citep{liu2019adaptive, liu2022entity}, which also uses reconstruction and contrastive learning \citep{refclip}; and temporal video grounding \citep{wstan, wstag}, where reconstruction generally outperforms contrastive methods \citep{scn, cnm, cpl}. The concurrent work \cite{kumar2025contextual} builds on phrase grounding. Our work \textbf{\textit{differentiates}} itself by integrating contrastive spatial grounding with reconstruction-based temporal grounding, enhancing contextual focus through isolation.

\noindent \textbf{Foundation Models for Referral Grounding:}  
Recent research has explored grounding via large vision foundation models (VFMs), which can be categorized into two types: 1) Large Vision Language Models (LVLMs) \citep{uniter-lvlm, uninext-lvlm, ofa-lvlm, Liu2023GroundingDM}, which combine representation learning with task-specific heads or are sequence-to-sequence decoders, and 2) Multimodal Large Language Models (MLLMs) \citep{sphinx-mllm, qwenvl-mllm, lenna-mllm, llavag-mllm, griffon-mllm}, which project vision features into pre-trained LLMs with task-specific heads. Both types address one or more tasks.
For our work, we focus on two key factors in selecting the VFM: 1) performance on image-referral grounding, where Grounding-DINO \citep{Liu2023GroundingDM} excels on RefCOCO+ \citep{refcoco} and RefCOCOg \citep{refcocog}, and 2) extensibility to WSTVG. MLLMs use language-centric causal generation for output, making end-features less suitable for spatio-temporal grounding. Grounding-DINO provides a more effective feature extraction. \textbf{\textit{In contrast}} to existing approaches, we extend Grounding-DINO, an image model, to the videos for STVG.

\noindent \textbf{Curriculum Learning} organizes task difficulty from simple to complex during training \citep{cur1, cur2}, mirroring the human cognitive learning processes. This approach has been applied across various tasks, including image classification \cite{img_cur}, object detection \cite{objdet_cur}, and temporal grounding \cite{cpl}, among others. To the best of our knowledge, our work is the first to apply a curriculum-based training strategy to WSTVG.

\section{Methodology}
\label{sec:method}

\noindent \textbf{Problem formulation} In WSTVG, the input consists of an untrimmed video $V=(v_{1}, v_{2}, ...v_{\textit{L}})$ with $L$ frames, along with a descriptive query caption $Q$ that specifies the primary subject and activity within the video. The task aims to output a spatio-temporal tubelet for the target subject, denoted as $A_R = \{a_{r}\}_{{t_{s}}}^{{t_{e}}}$, where $a_r$ represents the \textit{target} subject mentioned in the query, and $t_{s}$ and $t_{e}$ indicate the activity’s start and end timestamps. Under weak supervision, only video-level text annotations are available during training, without any detailed spatio-temporal labels for guidance.

\paragraph{Overview} Next, we describe our proposed approach. Firstly, we adapt G-DINO for STVG (Sec \ref{sec:wgdino}). We introduce a stronger baseline, TRG (Sec \ref{sec:trg}) over this naive adaptation. TRG struggles with comprehending dense spatio-temporal scenes and handling complex free-form queries. We propose STPro (Sec \ref{sec:stpro}) to alleviate these issue in two steps: Firstly (Sec. \ref{sec:tcl}), we breakdown complex queries into simpler sub-queries and train the model gradually over more difficult sub-queries to induce better temporal understanding in TRG. Secondly, to improve spatio-temporal scene understanding, in (Sec. \ref{sec:scl}) we develop a training paradigm that gradually increase scene complexity based on actor tubelets overlap to improve TRG's spatial grounding ability.

\subsection{Preliminaries: Referral VFMs}
\label{sec:wgdino}
Vision Foundation Models (VFMs) for image-based referral grounding take a textual query and an image as input to output the spatial location of the \textit{target subject}. In this work, we adapt VFMs for WSTVG, a video grounding task. Specifically, we utilize a text-based query $Q$ alongside video frames $I_f=\{V_{f}\}_{f=1}^{T}$, where $T$ represents the video duration. Since VFMs are inherently image-based while STVG requires spatio-temporal detection, we adapt Grounding DINO (G-DINO) \cite{Liu2023GroundingDM} for STVG task. This adaptation entails performing object detection on each frame. For spatio-temporal detections, we need bounding boxes connected across frames (\textit{tubelets}). To generate the tubelet from per-frame bounding boxes, we run a tracker algorithm \cite{Aharon2022BoTSORTRA} to get the tubelet $\mathcal{T}_{o_{k}}$ for each subject $k$, with $K$ indicating the total number of subjects in the video. This adapted model is termed weakly-supervised Grounding DINO (W-GDINO). To evaluate W-GDINO's efficacy, we consider the average confidence score for each tubelet and select the one with the highest score as W-GDINOs prediction. Despite the competitive results in Table \ref{tab:sota_weakly_hcstvg}, G-DINO's coarse-grained temporal alignment, which lacks semantic incorporation in tracking, and high confidence variance across tubelet frames (e.g., partial body, difficult poses, cut-scenes) limit its effectiveness for the task. To address this, we propose Tubelet Referral Grounding (TRG).

\begin{figure}[t!]
    \centering
    \includegraphics[width=\linewidth]{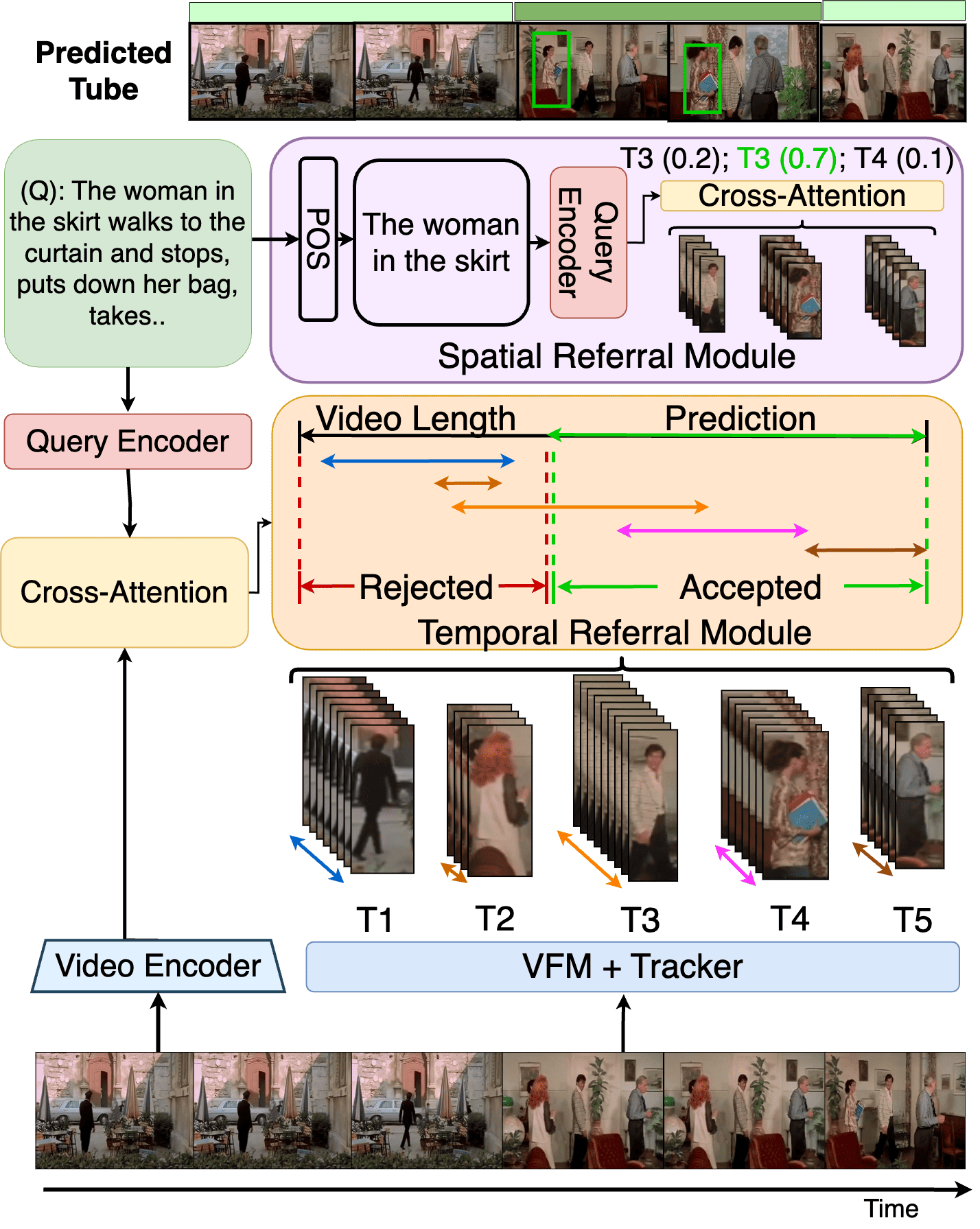} 
    \caption{\textbf{Overview of TRG:} TRG contains two grounding modules namely, TRM and SRM. TRM predicts the temporal action boundary via cross-attention between vision and query features. SRM grounds the correct referral subject tubelet from a set of TRM selected candidates having $tIoU(\mathcal{T}_{o_{k}}, Pred) > T_{filt}$. More details can be found in the supplementary.}
    \label{fig:trg}
\end{figure}

\begin{figure*}[t!]
    \centering
    \includegraphics[width=\linewidth]{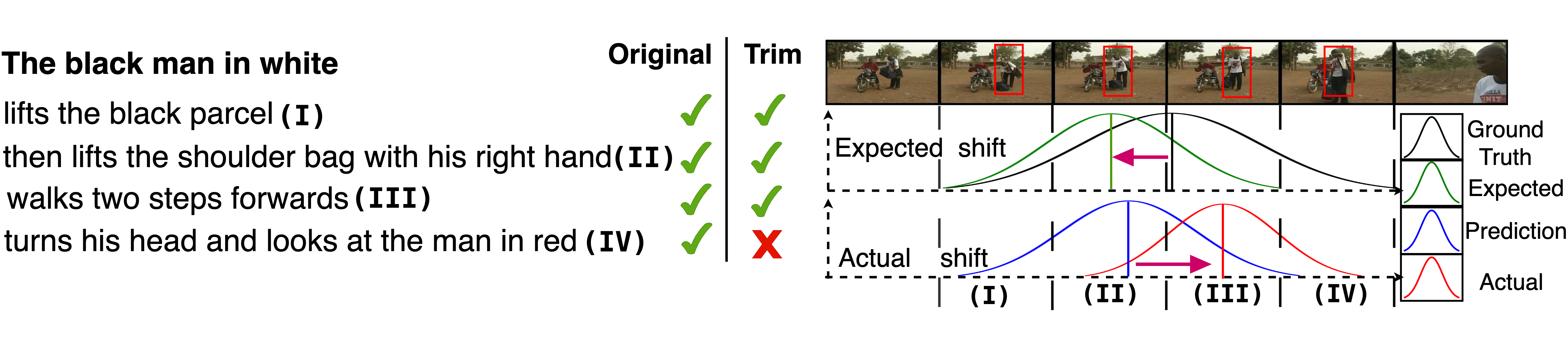}
    \caption{\textbf{Temporal Limitations of TRG:} When given a trimmed query (see trim in the figure), TRM misaligns temporal predictions, shifting right instead of left for missing final actions. This indicates it grasps temporal boundaries but lacks compositional understanding of individual actions, limiting generalization to novel sequences.}
    \label{fig:trg_failure_temporal}
\end{figure*}

\begin{figure}[t!]
    \centering
    \includegraphics[width=\linewidth]{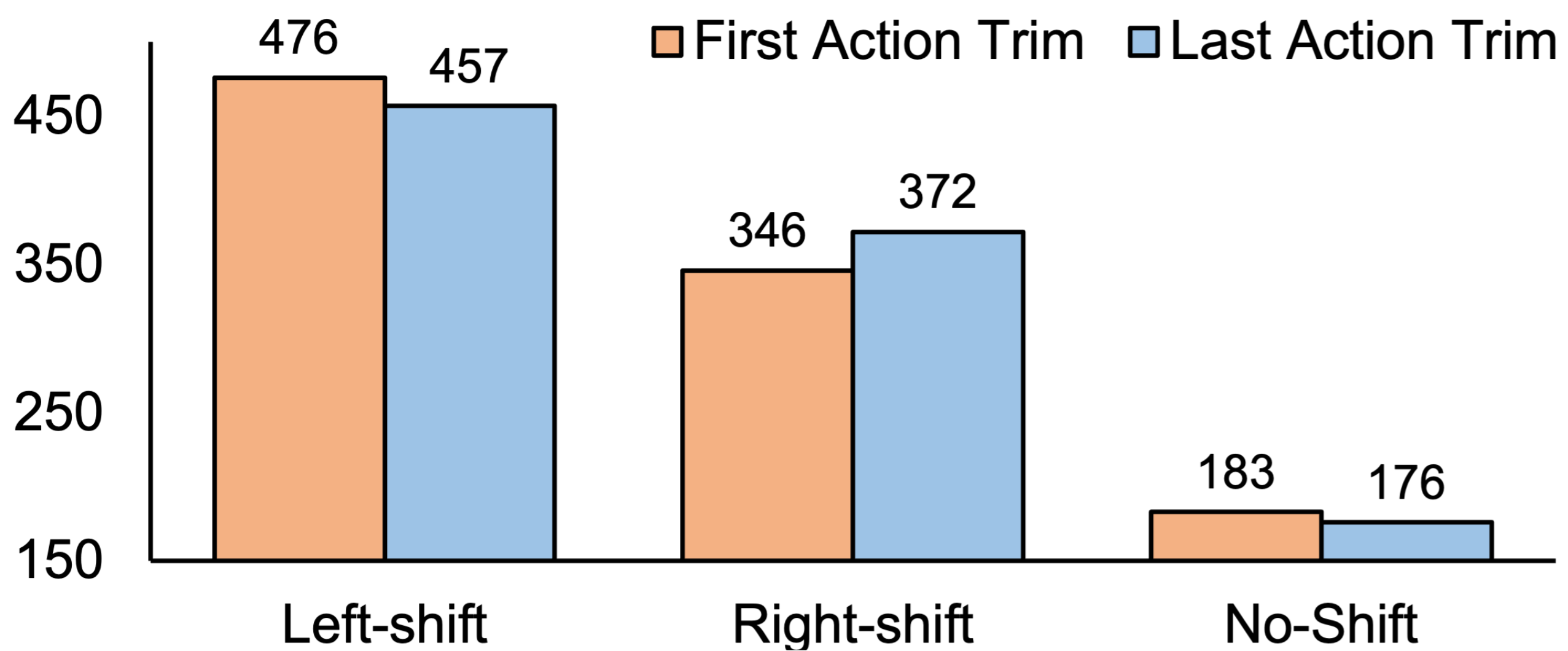}
    \caption{\textbf{Temporal-Shift Distribution of TRM:} Across HCSTVG-1, TRM struggles with temporal compositionality, failing to shift predictions correctly when queries are trimmed. Instead, it predominantly exhibits left/no-shift for first-action trimming and right/no-shift for last-action trimming, revealing its limitations.}
    \label{fig:trg_failure_temporal_plot}
\end{figure}

\subsection{Tubelet Referral Grounding (TRG)}
\label{sec:trg}
The TRG module adapts image-based foundational models for spatio-temporal detection by using contrastive training over tubelets, rather than frame-level detections. This approach effectively compresses a video into a single frame, to allow easy use of contrastive methods. For the STVG task, both temporal and spatial grounding are required. We introduce TRG as a two-stage process: the \textit{Temporal Referral Module} (TRM), which identifies the start and end timestamps of the target's actions, and the \textit{Spatial Referral Module} (SRM), which locates the target based on the textual query. TRM filters out irrelevant tubelets, while SRM grounds the correct tubelet within the temporal bounds. The module uses visual and query encoders following prior works \citep{liu2019adaptive, liu2022entity, refclip, cnm, cpl}.

\noindent \textbf{Visual encoder:} We extract object-level representations \( f_{o_k, t} = F_o(o_{k_t}) \in \mathbb{R}^{256} \) from G-DINO (based on DETR \citep{detr}), where \( F_o \) is the object encoder and \( o_{k_t} \) denotes the \( k \)-th detected subject at time \( t \). Detections are linked using a tracking algorithm \citep{Aharon2022BoTSORTRA} to form tubelets \( \mathcal{T}_{o_k} = \{o_{k_t}\}_{t=s}^{e} \), where \( s \) and \( e \) denote the subject’s first and last appearance. The tubelet features are given by \( \mathcal{F}_T = \{ \{ f_{o_k, t} \}_{t=s}^{e} \}_{k=1}^{K} \in \mathbb{R}^{K \times T_k \times 256} \), where \( T_k = e - s + 1 \) is the tubelet duration. Video features are extracted using a video encoder \( F_v \) (e.g., I3D \citep{Carreira2017QuoVA}) to obtain clip-level features \( f_c = F_v(\{V_t\}_{t=1}^{C}) \in \mathbb{R}^{C \times 1024} \), where \( C \) is the number of clips. 

\noindent \textbf{Query encoder:} The query $Q$ is first deconstructed to $Q_{w} = \{\mathtt{Linear}(\{w_{m}\}_{m=1}^{N})\} \in \mathbb{R}^{N\times 512}$ by linearly projecting pretrained GloVe embeddings ($w_i \in \mathbb{R}^{1 \times 300}$) for each query word. $N$ is the total query words. The word-level embeddings are enriched, to get  $Q_{e} = \mathtt{SA}(\mathtt{GRU}(Q_{w})) \in \mathbb{R}^{N\times 512}$ by a $\mathtt{GRU}$ followed by $\mathtt{SA}$, a single self-attention layer. Finally, we combine word-level features using attention-weighted flattening to get a single representative query feature  $Q_{f} \in \mathbb{R}^{1\times 512}$. 

\noindent 
\textbf{Temporal Referral Module (TRM)} identifies the temporal boundaries of the target subject's activity as specified in the textual query. Initial tubelet timestamps are often coarse, typically defined by the presence of the actor throughout the video, which complicates spatial localization and limits applicability to WSTVG. TRM refines this by filtering tubelets temporally, allowing the spatial module to focus on relevant video segments. Building on the effectiveness of reconstruction-based approaches for temporal grounding \citep{scn, deco, cnm, cpl}, we adopt \cite{cnm} as our baseline. It enforces semantic consistency between the target subject's actions and the textual query through a temporal grounding loss, $\mathcal{L}_t = - \sum_{m=1}^{N} log \mathcal{P}(q_w|f_{c}^{'}, \Tilde{q}_{[0:m-1]})$, where $q_w$ denotes the individual word feature.

\begin{figure}[t!]
    \centering
    \includegraphics[width=\linewidth]{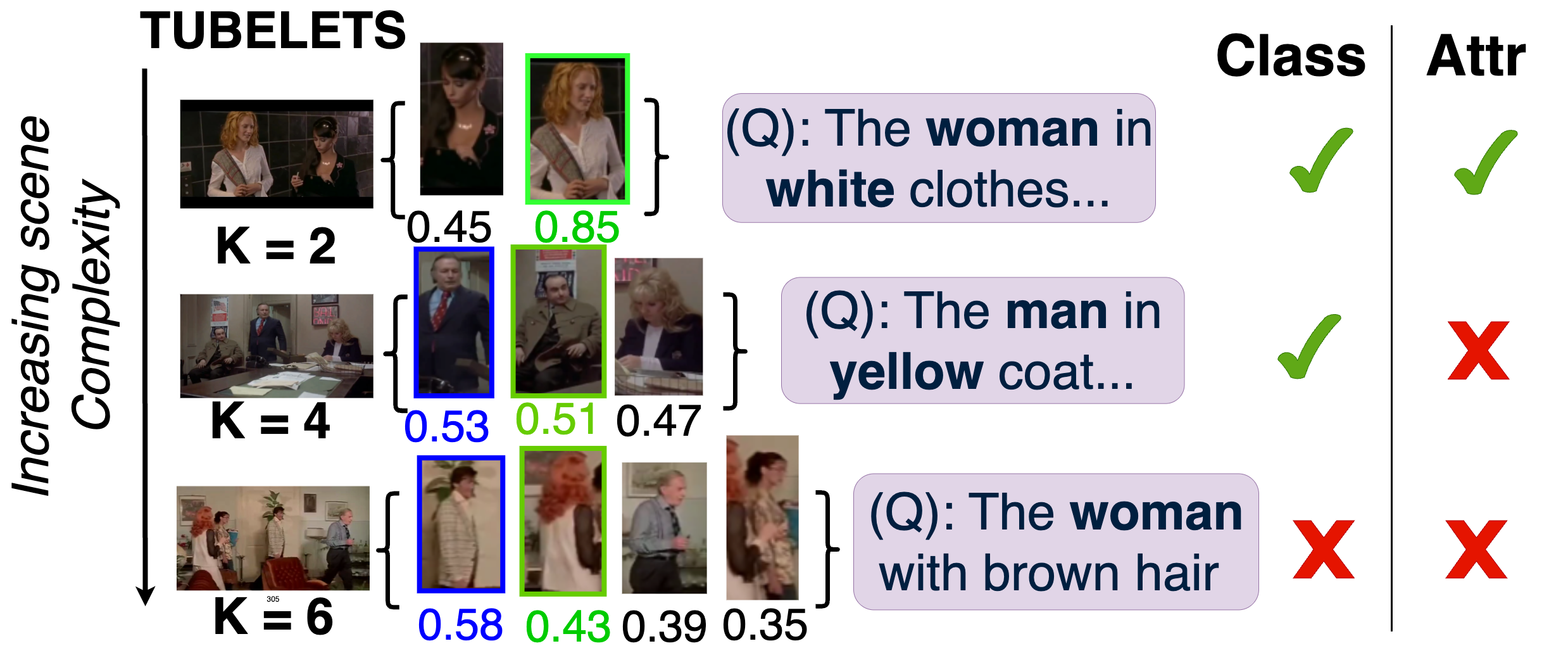}
    \caption{\textbf{Spatial limitations of TRG}: SRM performs well with fewer candidates but struggles with larger sets, misidentifying subject class and attributes, indicating inadequate cue learning during training. \textcolor{blue}{Blue} denotes predictions, \textcolor{green}{green} denotes ground truth.}
    \label{fig:trg_failure_spatial}
\end{figure}

\textbf{Spatial Referral Module (SRM):} localizes the target subject from a set of candidate subjects within the temporal bounds predicted by TRM. We employ multimodal contrastive loss to align tubelet features and textual queries, building on \cite{datta2019align2ground, gupta2020contrastive, wang-etal-2020-maf, wang2021improving}. Adapting \cite{refclip} for STVG, we compute the similarity between each candidate tubelet and the textual query, averaging the per-frame detection features within a tubelet for improved spatial localization (see Table \ref{tab:compare_frame_tubelet}). To address background noise in the textual query, we extract noun phrases ($\mathtt{POS}$) focusing on the target subject and its attributes, enhancing the model's discriminative ability (shown in Table \ref{tab:abla_all}). The spatial loss ($\mathcal{L}_s$), is defined as
\noindent

{\small
\begin{equation}
\mathcal{L}_s = -\log \frac{\exp\left(\scriptstyle{\text{Sim}_{avg}(\mathcal{F}_{T_0}^i, Q_{f}^i) / \tau}\right)}
{\sum_{n=0}^N \sum_{j=0}^M \mathbb{1}_{(i=j \land n \neq 0)} \exp\left(\scriptstyle{\text{Sim}_{avg}(\mathcal{F}_{T_n}^j, Q_{f}^i)/ \tau} \right)}
\label{eq:region_att}
\end{equation}
}
\begin{equation}
\text{Sim}_{avg}(\mathcal{F}_{T_k}, Q_{f}) = \frac{1}{K} \sum_{i=1}^{K} \text{sim}\left(\mathcal{\text{f}}_{o_k,i}, Q_{f}\right),
\end{equation}

where $N$ and $M$ are the number of contrastive negatives in the sample and in the batch respectively.

\noindent \textbf{Limitations of TRG:} 
While TRG establishes a strong baseline for STVG, it faces challenges in the comprehension of complex spatio-temporal scenes (shown in Figs. \ref{fig:trg_failure_temporal_plot} and \ref{fig:trg_failure_spatial}). Temporally, TRG fails to understand individual actions and their composition (Figs. \ref{fig:trg_failure_temporal} and \ref{fig:trg_failure_temporal_plot}). Spatially, Fig. \ref{fig:trg_failure_spatial} reveals that as scene complexity increases, the model's confidence and ability to ground the target actor decreases. This suggests that TRG struggles with complex free-form queries and dense scene understanding, particularly due to the inconsistent presence of multiple actors. To address these limitations, we propose STPro.

\subsection{Spatio-Temporal Progressive Learning (STPro)}
\label{sec:stpro}
\begin{figure}[t!]
    \centering
    \includegraphics[width=0.95\linewidth]{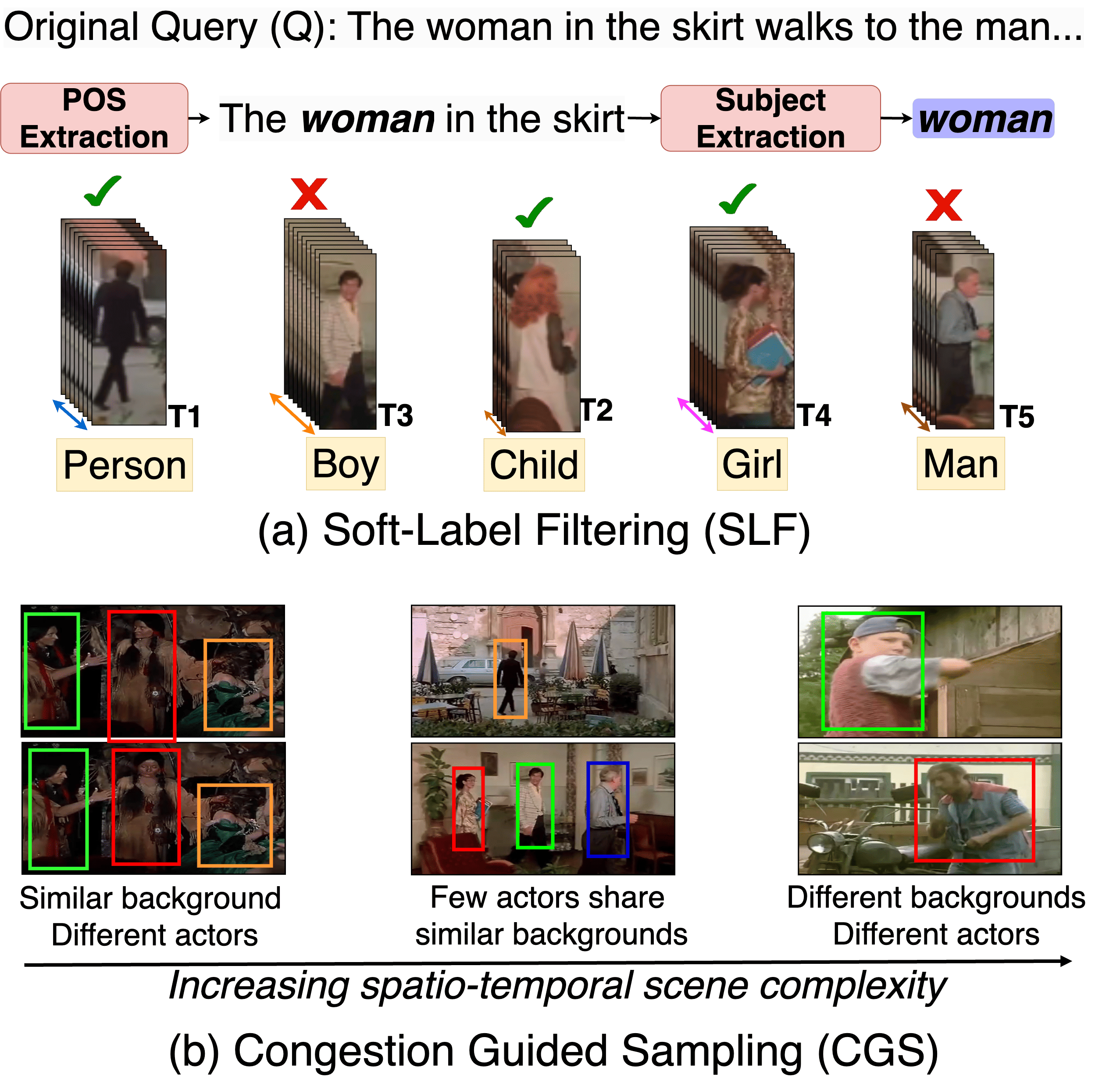}
    \caption{\textbf{STPro (CG-SCL) Overview:} CG-SCL comprises of Soft-Label Filtering (SLF) and Congestion Guided Sampling (CGS). (a) SLF filters tubelets using G-DINO’s zero-shot labels, selecting those matching the caption’s subject type while also including generic tubelets such as \textit{person}. (b) CGS organizes training from high to low pairwise tubelet $tIoU$: simple cases with similar backgrounds to complex scenes with diverse backgrounds, ensuring that the model first learns actor-specific features before handling background variability. T denotes tubelets.}
    \label{fig:stpro_cgscl_overview}
\end{figure}

STPro consists of two components to address the challenges and enhance the model's referral capabilities: Sub-Actions Temporal Curriculum Learning (SA-TCL) (Sec. \ref{sec:tcl}) and Congestion-Guided Spatial Curriculum Learning (CG-SCL) (Sec. \ref{sec:scl}). SA-TCL focuses solely on the temporal domain. It improves action composition understanding by progressively learning the constituent actions of compound queries in a staged temporal curriculum. CG-SCL gradually increases scene complexity, improving tubelet-query alignment for better spatial understanding. It uses temporal tubelet congestion as a signal for sample hardness. Fig. \ref{fig:stpro_satcl_overview} and Fig. \ref{fig:stpro_cgscl_overview} give an overview of STPro.

\subsubsection{Sub-Actions Temporal Curriculum Learning}
\label{sec:tcl}
\noindent \textbf{Motivation:} 
Effective temporal localization in STVG requires accurate identification of both the target subject and the actions they perform, as multiple actors may share similar actions over overlapping time spans. A strong understanding of action composition improves the model's adaptability, enabling it to generalize to novel action sequences by recombining learned action subsets. Existing temporal grounding methods \cite{wstag, wstan, scn, cnm, cpl} lack referral capabilities, localizing all actions without distinguishing actor-specific actions. 

To evaluate the action compositionality of TRM, we created test samples by trimming the first and last actions from the query, keeping the order of remaining actions intact. For example, for the query: \textit{“The black man in white \underline{lifts} the black parcel, then \underline{lifts} the shoulder bag with his right hand, \underline{walks} two steps forward, \underline{turns} his head and looks at the man in red”}, the start-trimmed query becomes \textit{“The black man \underline{lifts} the shoulder bag with his right hand, \underline{walks} two steps forward, \underline{turns} his head and looks at the man in red”}, and the end-trimmed query becomes \textit{“The black man in white \underline{lifts} the black parcel, then \underline{lifts} the shoulder bag with his right hand, \underline{walks} two steps forward”}. We analyze how TRM handles these modified queries (Fig. \ref{fig:trg_failure_temporal}).

\begin{figure}[t!]
    \centering
    \includegraphics[width=\linewidth]{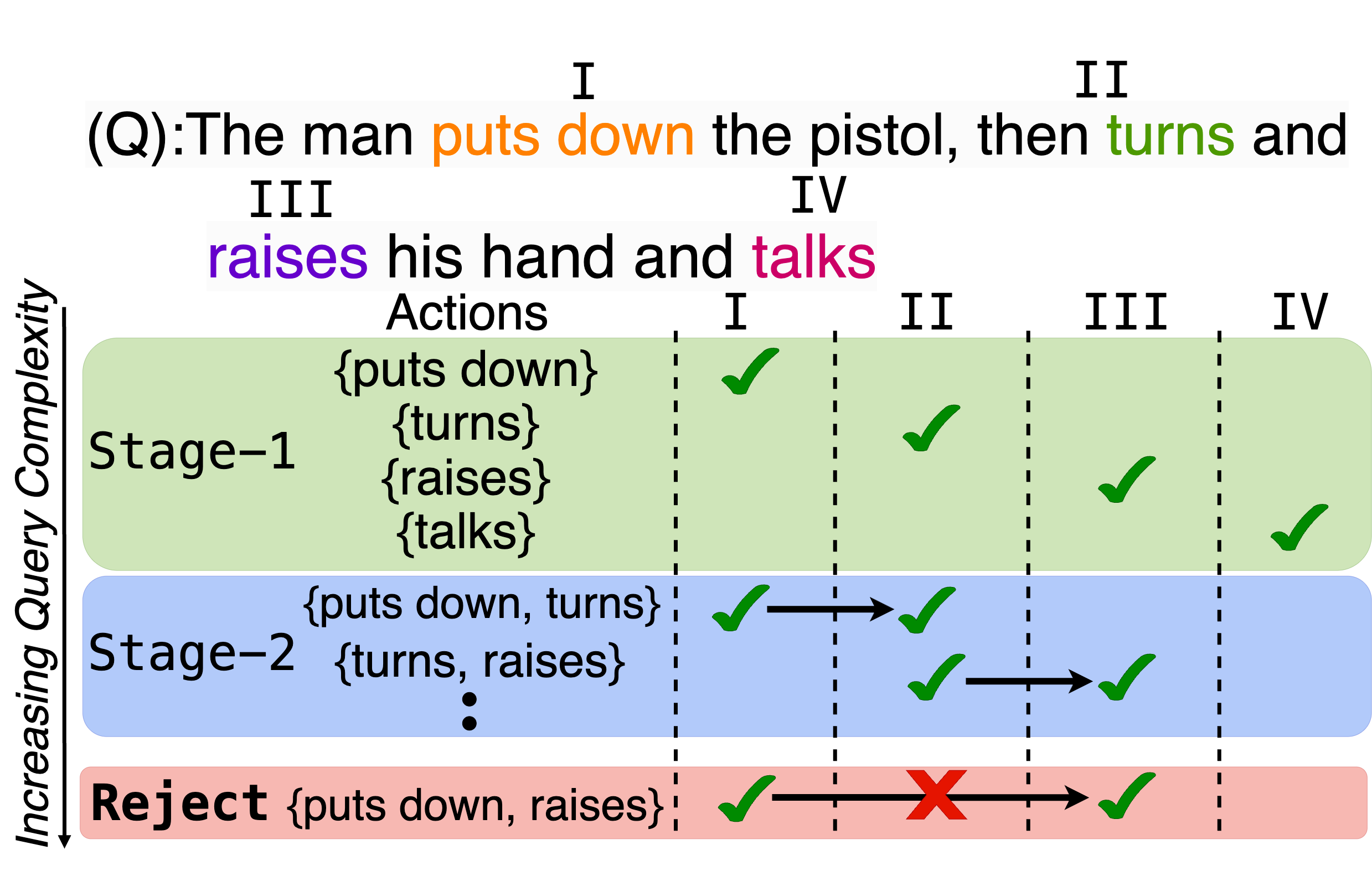}
    \caption{\textbf{STPro (SA-TCL) Overview:} SA-TCL enhances compositional understanding by iteratively grounding sub-part combinations of an original query. The stages of SA-TCL progress from single actions (stage-1) to action-pairs (stage-2) and so on. Sub-part combinations in which actions are non-continuous are rejected. }
    \label{fig:stpro_satcl_overview}
\end{figure}

The model's temporal prediction is ideally adjusted based on the trimmed actions: shifting the start time in the case of start-trimming and shifting the end time in the case of end-trimming. We compare the shift of the temporal midpoint in all three cases, expecting a right shift for start-trimming and a left shift for end-trimming. Our results show that in most cases, the model fails to adjust appropriately, \textbf{incorrectly shifting the temporal boundary 65\% of the time for start-trimming and 54\%} for end-trimming. This indicates that while the model captures some subtle query changes, it largely fails to understand the semantics of action composition.

To mitigate this limitation, we propose \textbf{SA-TCL} to learn action composition via curriculum learning. We utilize a Large Language model (LLM) (for e.g. GPT 3.5-turbo) to break down queries into all constituent ordered action combinations i.e. we do not skip over any actions. We increase the complexity of training by increasing the number of actions in each successive stage starting from single actions. 
Consider the original query - \textit{The man puts down the pistol, then turns and raises his hand and talks.}  The first stage comprises of captions containing only individual actions, such as: \textit{The man \underline{puts} down the pistol. The  man \underline{raises} his hand.} The second stage comprises coupled and ordered actions: \textit{The man \underline{puts} down the pistol, then \underline{turns}.} \textit{The man \underline{turns} and \underline{raises} his hand.}
and so on for successive stages (shown in Fig. \ref{fig:stpro_satcl_overview}). Further details on the prompt and training stages are in the supplementary.

\subsubsection{Congestion Guided Spatial Curriculum Learning}
\label{sec:scl}

\noindent \textbf{Motivation:} Given the $\mathtt{POS}$-extracted textual query, SRM should ideally distinguish between actors throughout the video without being influenced by temporal cues (e.g., action sequences). However, as scene complexity increases and similar tubelets (e.g., multiple instances of $\mathtt{man}$) appear, the model's ability to correctly distinguish the target actor diminishes (Fig. \ref{fig:trg_failure_spatial}). This suggests $\mathtt{POS}$ benefits are underutilized since: \textit{First}, the model must simultaneously distinguish actor types and attributes, and \textit{Second}, sparse temporal overlap among tubelets makes fine-grained feature extraction harder, as varying backgrounds influence the feature set.

To overcome these challenges, we propose: 1) \textbf{Soft-Label Filtering (SLF)}, which helps the model learn actor-type correspondence and focus on actor attributes to distinguish similar actors in complex spatial scenarios, and 2) \textbf{Congestion Guided Sampling (CGS)}, a curriculum learning approach that gradually improves actor and attribute selection by promoting slower feature divergence.

\noindent \textbf{Soft-Label Filtering (SLF):} SLF leverages the zero-shot capabilities of G-DINO to filter tubelets by their dominant soft labels (Refer Fig \ref{fig:stpro_cgscl_overview} (a)). Given detection labels \( \{l_1, \dots, l_n\} \) for each frame of tubelet \( T_k \), majority label (mode) determines tubelet \textit{type}. Tubelets are selected if their \textit{type} matches the \textit{subject type} from caption. Additionally, we always include tubelets with high variability in soft labels due to noisy frames (e.g., partial bodies) for training. If the extracted subject is non-specific (e.g., "person"), we include all tubelets for training, regardless of their dominant soft-label.

\noindent \textbf{Congestion Guided Sampling (CGS):} CGS calculates the average pairwise temporal IoU across all tubelets in a video, which for ($N$) tubelets is calculated as: $\frac{1}{\binom{N}{2}} \sum_{i=1}^{N-1} \sum_{j=i+1}^{N} \text{tIoU}(T_i, T_j)$. The curriculum progresses from high to low $tIoU$: simple scenes with similar backgrounds to more complex scenes with dynamic backgrounds (Refer Fig. \ref{fig:stpro_cgscl_overview} (b)). This gradual approach allows the model to focus on the actor features first, then on both the actor and the background features.

\section{Experiment Details}
\label{sec:exp_details}
\noindent \textbf{Datasets:} We evaluate performance on three benchmark datasets: VidSTG \citep{vidstg}, HCSTVG-v1 \citep{hcstvg}, and HCSTVG-v2 \citep{hcstvg}. VidSTG contains 99,943 video-sentence pairs (44,808 declarative, 55,135 interrogative) across 10,303 videos and 80 object categories, split into 80,684 training, 8,956 validation, and 10,303 test pairs, with 5,436, 602, and 732 unique videos per subset. HCSTVG-v1 consists of 4,500 training and 1,160 test videos, focusing on human attributes and actions. HCSTVG-v2 expands upon HCSTVG-v1 with 16,544 videos, divided into 10,131 training, 2,000 validation, and 4,413 testing videos. Since the HCSTVG-v2 test set is unavailable, we report results on the validation set, following prior studies \citep{Yang2022TubeDETRSV, clb, cg-stvg}. 

\noindent \textbf{Implementation details:} We divide our approach into three components:  (a) \textbf{SRM} - We apply G-DINO \citep{Liu2023GroundingDM} with a 0.4 threshold for phrase and box detection, running detection every 5th frame and extracting decoder-layer features. Tube generation is managed via BoTSORT \citep{Aharon2022BoTSORTRA}. (b) \textbf{TRM} - We sample 32 indexed frames for tubelet features and extract video-level features using I3D \citep{Carreira2017QuoVA}. (c) \textbf{SA-TCL \& CG-SCL} - Using GPT-3.5, we extract parts-of-speech ($\mathtt{POS}$) tags and sub-action phrases. CG-SCL follows five training stages, decreasing temporal IoU from 1.0 to 0.0 in steps of 0.2, while SA-TCL progressively increases action complexity over four stages, each trained for 50 epochs. Additional details are in the supplementary.

\noindent \textbf{Inference:} Temporal localization module (TRM) predicts the start ($t_s$) and end ($t_e$) time stamps for the target actor's actions. The tubelet with highest attention from the spatial localization module (SRM) within these temporal bounds is the predicted tubelet $\hat{a}$.

\begin{table}[t]
	\centering
	\renewcommand{\arraystretch}{1.06}
	\scalebox{0.8}{
		\begin{tabular}{r cccc}
				\rowcolor{mygray} 
				\specialrule{1.5pt}{0pt}{0pt}
    \cellcolor{mygray}  & \multicolumn{4}{c}{ \cellcolor{mygray} HCSTVG - v1} \\
        \rowcolor{mygray} Methods & vIoU & vIoU@0.3 &  vIoU@0.5 \\
    \hline\hline
    \textit{Two-stage pipelines} \\ \hline
     GroundeR \textcolor{lightgray}{\scriptsize{[ECCV16]}}~\citep{grounder}+LCNet \textcolor{lightgray}{\scriptsize{[IEEE17]}}~\citep{lcnet} &  4.17 & 3.28 & 1.05  \\
    MATN \textcolor{lightgray}{\scriptsize{[CVPR18]}}~\citep{matn}+LCNet \textcolor{lightgray}{\scriptsize{[IEEE17]}}~\citep{lcnet} & 4.41 & 3.53 & 1.12  \\
    GroundeR \textcolor{lightgray}{\scriptsize{[ECCV16]}}~\citep{grounder}+CPL \textcolor{lightgray}{\scriptsize{[CVPR22]}}~\citep{cpl} &  5.23 & 4.18 & 1.25  \\
    RAIR \textcolor{lightgray}{\scriptsize{[CVPR21]}}~\citep{rair}+CPL \textcolor{lightgray}{\scriptsize{[CVPR22]}}~\citep{cpl} & 6.88 & 4.87 & 1.36  \\ \hline
    \textit{Single-stage pipelines} \\ \hline
    WSSTG \textcolor{lightgray}{\scriptsize{[ACL19]}}~\citep{chen-etal-2019-weakly} & 6.52 & 4.54 & 1.27  \\
    AWGU \textcolor{lightgray}{\scriptsize{[ACMMM20]}}~\citep{Chen2020ActivitydrivenWS} & 8.20 & 4.48 & 0.78 \\
    Vis-CTX \textcolor{lightgray}{\scriptsize{[CVPR19]}}~\citep{nafae} & 9.76 & 6.81 &  1.03  \\
    WINNER \textcolor{lightgray}{\scriptsize{[CVPR23]}}~\citep{Li_2023_CVPR} & 14.20 & 17.24 & \underline{6.12} \\ 
    VCMA \textcolor{lightgray}{\scriptsize{[ECCV24]}}~\citep{vcma} & \underline{14.64} & \underline{18.60} & 5.75 \\ \hline
     W-GDINO (Ours-Baseline)  & 9.04 & 11.56 & 4.57 \\
     STPro (Ours) &  \textbf{17.56} & \textbf{26.98} & \textbf{12.93}  \\
     & \improve{~+2.92} &\improve{~+8.38} & \improve{~+6.81}   \\
     \specialrule{1.5pt}{0pt}{0pt}
\cellcolor{mygray}  & \multicolumn{4}{c}{ \cellcolor{mygray} HCSTVG - v2} \\
      \hline
     W-GDINO (Ours-Baseline) & \underline{9.85}  & \underline{13.30} & \underline{5.63} \\ 
     STPro (Ours)& \textbf{19.99} & \textbf{31.70} & \textbf{14.55} \\
     &  \improve{~+10.14} & \improve{~+18.40} & \improve{~+8.92} \\
     \specialrule{1.5pt}{0pt}{0pt}
		\end{tabular}}
  \caption{Comparison with existing state-of-the-art weakly-supervised methods on HCSTVG-v1 and v2 datasets. \textbf{Bold} denotes best and \underline{underline} denotes second best.}
	\label{tab:sota_weakly_hcstvg}
\end{table}

\begin{table*}[t]
	\centering
	\renewcommand{\arraystretch}{1.06}
	\scalebox{0.83}{
		\begin{tabular}{r ccc ccc}
			\specialrule{1.5pt}{0pt}{0pt}
			\rowcolor{mygray} 
			\cellcolor{mygray} & \multicolumn{3}{c}{ \cellcolor{mygray} Declarative Sentences} & \multicolumn{3}{c}{ \cellcolor{mygray}Interrogative Sentences} \\ 
			\rowcolor{mygray} 
			\multirow{-2}{*}{\cellcolor{mygray} Methods}  & m\_vIoU & vIoU@0.3 &  vIoU@0.5  & m\_vIoU & vIoU@0.3 &  vIoU@0.5  \\
			\hline
			\hline
    \textit{Two-stage pipelines} \\ \hline
    GroundeR\textcolor{lightgray}{\scriptsize{[ECCV16]}}~\citep{grounder}+LCNet  \textcolor{lightgray}{\scriptsize{[IEEE17]}}~\citep{lcnet} &  7.85 & 7.96 & 3.02 &  6.43 & 6.58 & 2.92 \\
    MATN\textcolor{lightgray}{\scriptsize{[CVPR18]}}~\citep{matn}+LCNet \textcolor{lightgray}{\scriptsize{[IEEE17]}}~\citep{lcnet} &  8.16 & 8.03 & 3.59 &  6.97 & 6.64 & 3.05 \\
    GroundeR\textcolor{lightgray}{\scriptsize{[ECCV16]}}~\citep{grounder}+CPL\textcolor{lightgray}{\scriptsize{[CVPR22]}}~\citep{cpl} &  8.28 & 8.35 & 3.68 &  7.16 & 7.28 & 3.23 \\
    RAIR \textcolor{lightgray}{\scriptsize{[CVPR21]}}~\citep{rair}+CPL \textcolor{lightgray}{\scriptsize{[CVPR22]}}~\citep{cpl} & 8.67 & 8.72 & 4.01 & 7.68 & 7.71 & 3.58 \\ \hline
    \textit{Single-stage pipelines} \\ \hline
    WSSTG \textcolor{lightgray}{\scriptsize{[ACL19]}}~\citep{chen-etal-2019-weakly} & 8.85 & 8.52 & 3.87 & 7.12 & 6.87 & 2.96 \\
    AWGU \textcolor{lightgray}{\scriptsize{[ACMM20]}}~\citep{Chen2020ActivitydrivenWS} & 8.96 & 7.86 & 3.10 & 8.57 & 6.84 & 2.88 \\
    Vis-CTX \textcolor{lightgray}{\scriptsize{[CVPR19]}}~\citep{nafae} &  9.34 & 7.32 & 3.34 & 8.69 & 7.18 & 2.91   \\
    WINNER \textcolor{lightgray}{\scriptsize{[CVPR23]}}~\citep{Li_2023_CVPR} & 11.62 & 14.12 & 7.40 & 10.23 & 11.96 & 5.46  \\ 
    VCMA  \textcolor{lightgray}{\scriptsize{[ECCV24]}}~\citep{vcma} &  \underline{14.45} & \underline{18.57} & \underline{8.76} &  \textbf{13.25} & \textbf{16.74} & \underline{7.66} \\
    \hline
    W-GDINO (Ours-Baseline)  & 10.69 & 13.02 & 7.83 & 9.87 &  12.16 & 6.71  \\
     STPro (Ours)  & \textbf{15.52} & \textbf{19.39} & \textbf{12.69}  & \underline{12.56} & \underline{14.95} & \textbf{9.29}\\
     & \improve{~+1.07} & \improve{~+0.82} & \improve{~+3.93}  & \improve{~-0.69} & \improve{~-1.79} & \improve{~+1.63}\\
    	\specialrule{1.5pt}{0pt}{0pt}
	\end{tabular}}
 \caption{Comparison with existing state-of-the-art weakly-supervised methods on VidSTG dataset. \textbf{Bold} denotes best and \underline{underline} denotes second best.}
	\label{tab:sota_weakly_vidstg}
\end{table*}

\noindent \textbf{Evaluation Metrics: } We report performance using metrics established in previous studies \citep{Yang2022TubeDETRSV, Jin2022EmbracingCA}: mean average spatio-temporal localization (m\_vIoU) and mean temporal localization (m\_tIoU). m\_vIoU and m\_tIoU are computed as $\frac{1}{|S_{u}|} \sum_{t\in S_{i}}$IoU$(\hat{b_{t}}, b_{t})$ and $\frac{|S_{i}|}{|S_u|}$, where $S_{i}$ and $S_{u}$ denote the intersection and union, respectively, between the predicted and ground truth timestamps. IoU$(\hat{b}_t, b_t)$ represents the spatial overlap between the predicted bounding box $\hat{b}_t$ and the ground truth box $b_t$ at frame $t$. Additionally, vIoU@R indicates the proportion of samples with a mean vIoU greater than R, and we present results for R values of 0.3 and 0.5, consistent with prior work \citep{Yang2022TubeDETRSV, Li_2023_CVPR}.

\subsection{Results and comparisons}
\noindent \textbf{\textit{Comparison with two-stage weakly-supervised}} Tables \ref{tab:sota_weakly_hcstvg} and \ref{tab:sota_weakly_vidstg} shows that our approach outperforms previous weakly-supervised approaches on all metrics against two-stage approaches on both HCSTVG-v1 and VidSTG. Looking into HCSTVG-v1 dataset, we outperform RAIR\cite{rair}+CPL\cite{cpl}, the best two-stage approach by a margin of 10.68\%, 22.11\% and 11.57\% on mean vIoU, vIoU@0.3 and vIoU@0.5 respectively. On VidSTG dataset (Table \ref{tab:sota_weakly_vidstg}), STPro outperforms RAIR\cite{rair}+CPL\cite{cpl} by 6.85\% (declarative) and 4.88\% (interrogative) on mean vIoU.

\noindent \textbf{\textit{Comparison with single-stage weakly-supervised}} STPro outperforms previous weakly-supervised methods on most metrics. On HCSTVG-v1, it surpasses SoTa VCMA by 2.92\%, 8.38\%, and 6.81\% in mean vIoU, vIoU@0.3, and vIoU@0.5, respectively. Against the foundation model baseline W-GDINO, STPro improves mean vIoU and tIoU by 8.52\% and 12.59\% on HCSTVG-v1, and 10.14\% and 15.66\% on HCSTVG-v2. This confirms that while foundation model features are richer, our method significantly enhances performance. VidSTG remains highly challenging, with gains under 2\% in recent years. On VidSTG (Table \ref{tab:sota_weakly_vidstg}), we achieve competitive results on interrogative subsets and surpass prior work on declarative ones.

\subsection{Ablation study}
We study the effectiveness of different components of STPro on HCSTVG-v1.

\begin{table}[t]
	\centering
\renewcommand{\arraystretch}{1.1}
\scalebox{0.8}{
\begin{tabular}{ccc cccc}
    \rowcolor{mygray} 
    \specialrule{1.5pt}{0pt}{0pt}
    SRM & TRM  & $\texttt{\textbf{POS}}$ & m\_tIoU & m\_vIoU & vIoU@0.3 &  vIoU@0.5  \\ 
    \hline\hline
          &&& 17.97 & 9.04 & 11.56 & 4.57\\
        \checkmark & & & 20.08 & 9.99 & 12.50 & 5.17 \\
        \checkmark & & \checkmark & 20.15 & 10.40 & 12.67 & 5.01 \\
        \checkmark & \checkmark  & \checkmark & \textbf{29.40} & \textbf{14.81} & \textbf{21.81} & \textbf{10.26} \\
    \specialrule{1.5pt}{0pt}{0pt}
    TRG & CG-SCL & SA-TCL  & m\_tIoU & m\_vIoU & vIoU@0.3 &  vIoU@0.5  \\ 
    \hline\hline 
    &&& 17.97 & 9.04 & 11.56 & 4.57\\
\checkmark  && & 29.40 & 14.81 & 21.81 & 10.26 \\
\checkmark & & \checkmark & \textbf{30.67} & 15.31 & 22.84 & 10.34 \\
\checkmark & \checkmark && 29.35 & 17.15 & 26.55 & 12.16 \\
\checkmark & \checkmark & \checkmark & 30.56 & \textbf{17.56} & \textbf{26.98} & \textbf{12.93}\\
&&&\improve{~+12.59} & \improve{~+8.52} & \improve{~+15.42} & \improve{~+8.36} \\
 \specialrule{1.5pt}{0pt}{0pt}
\end{tabular}}
  \caption{\textbf{Ablation study} on sub-modules of TRG module(upper) and different components of STPro (lower).}
    \label{tab:abla_all}
\end{table}

\begin{figure*}[t!]
    \centering
    \includegraphics[width=0.95\linewidth]{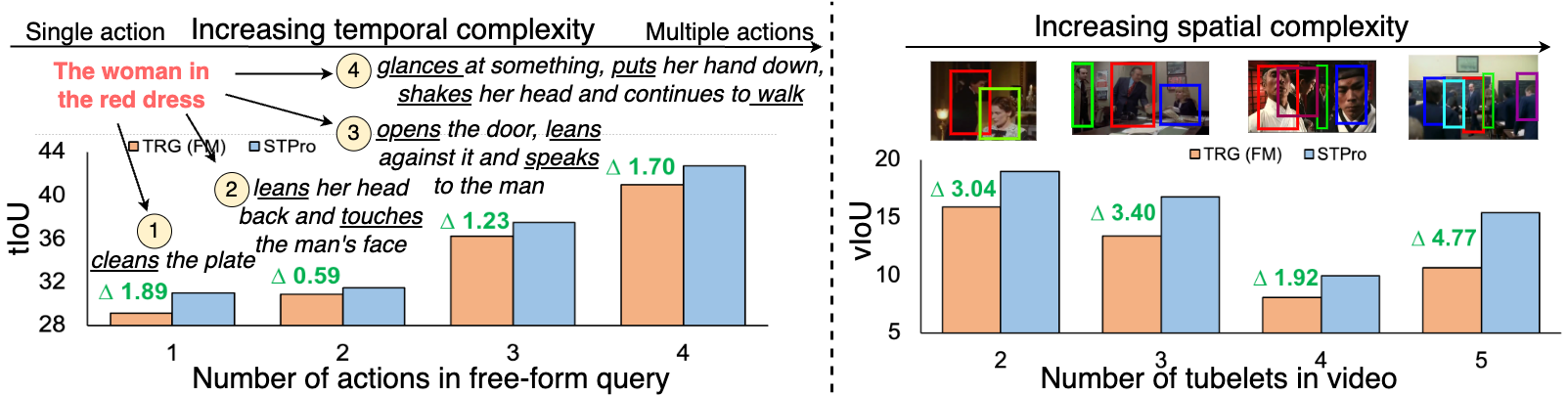} 
    \caption{\textbf{Qualitative Analysis:} Comparison of performance between TRG and STPro with increasing complex query for temporal grounding (left) and increasing scene complexity (number of tubelets) for spatial grounding (right).}
    \label{fig:qualitative}
\end{figure*}

\noindent \textbf{Effectiveness of TRG sub-modules} Table \ref{tab:abla_all} analyzes TRG components. SRM improves mean vIoU by 1\% over W-GDINO, proving its spatial discriminative ability. Using $\mathtt{POS}$ tags instead of full captions boosts mean vIoU by 0.41\%, supporting the hypothesis that $\mathtt{POS}$ tags help the model attend to the target subject and attributes. Adding the temporal module TRM further increases mean vIoU by 4.41\% and mean tIoU by 9.25\%, demonstrating its effectiveness in predicting accurate timestamps.

\noindent \textbf{Impact of SA-TCL and CG-SCL:} Table \ref{tab:abla_all} examines CG-SCL and SA-TCL. TRM outperforms W-GDINO across all metrics. Adding SA-TCL boosts mean vIoU by 0.6\% over TRM and 6.27\% over W-GDINO, while improving mean tIoU by 1.47\% on HCSTVG-v1. This highlights the benefit of incorporating action composition in curriculum learning for temporal grounding. CG-SCL, combined with TRG, raises mean vIoU by 2.34\%, showing that increasing tubelet congestion enhances spatial discrimination. Using both modules together improves mean vIoU by 2.75\% over TRG and 8.52\% over W-GDINO, proving their orthogonal contributions to spatio-temporal grounding. Notably, STPro excels on harder metrics, achieving 2× and 3× gains over W-GDINO at vIoU@0.3 and vIoU@0.5, respectively.

\begin{table}[t]
	\centering
    \renewcommand{\arraystretch}{1.1}
    \scalebox{0.8}{
        \begin{tabular}{l cccc}
        \rowcolor{mygray} 
        \specialrule{1.5pt}{0pt}{0pt}
        \rowcolor{mygray}\multicolumn{5}{c}{ \cellcolor{mygray} Temporal Curriculum Learning} \\
        \rowcolor{mygray}Methods & tIoU &  tIoU@0.1 & tIoU@0.3 &  tIoU@0.5  \\ 
        \hline\hline
        TRM & 32.96 & 77.65 & 56.95 & 23.55 \\
        Beta-Increment & 33.35 & 77.48 & 57.29 & 23.30\\
        Self-Advised & 33.41 & 78.26 & 57.12 & 23.73\\
        Sub-Actions (Ours) & \textbf{33.92} & \textbf{78.77}& \textbf{57.72} & \textbf{24.68} \\ 
    \specialrule{1.5pt}{0pt}{0pt}
    \rowcolor{mygray}\multicolumn{5}{c}{ \cellcolor{mygray} Spatial Curriculum Learning} \\
    \rowcolor{mygray}Methods & m\_vIoU &  vIoU@0.1 & vIoU@0.3 &  vIoU@0.5  \\ 
    \hline\hline
    SRM + $\mathtt{POS}$ & 10.40 & 30.69 & 12.67 & 5.00 \\
    + SLF & 11.92 &  33.71 & 15.26 & 6.72\\
    + CGS & 11.01 & 32.41 & 13.36 & 5.43\\
    + (SLF + CGS) & \textbf{12.61} & \textbf{35.52} & \textbf{16.38} & \textbf{7.33}\\
    \specialrule{1.5pt}{0pt}{0pt}
    \end{tabular}}
  \caption{Comparison against different TCL (upper) and SCL (lower) approaches. TCL and SCL shows performance on standalone temporal and spatial metrics. }
  \label{tab:discuss_tcl_scl}
\end{table}

\noindent \textbf{Qualitative Analysis} Fig. \ref{fig:qualitative} shows the effectiveness of our approach along both spatial and temporal dimensions.

\section{Discussion and analysis}

\begin{table}
    \centering
    \renewcommand{\arraystretch}{1.1}
    \scalebox{0.77}{
        \begin{tabular}{c | cc | cc  ccc}
            \specialrule{1.5pt}{0pt}{0pt}
            \rowcolor{mygray} 
   \rowcolor{mygray} Approach & Frame & Tubelet &  m\_vIoU &vIoU@0.1&vIoU@0.3&vIoU@0.5 \\ \hline\hline
    TRG & \checkmark & &14.79 & 48.62 & 16.72 & 4.22\\
    TRG & & \checkmark & 14.81 & 37.41& 21.81 & 10.26\\ \hline
    STPro & \checkmark &&16.64 & 53.88 & 19.48 & 4.48 \\
    STPro & & \checkmark & 17.56 & 41.90 & 26.98 & 12.93 \\ 
    \specialrule{1.5pt}{0pt}{0pt}
			\end{tabular}}
    \caption{Comparison between frames and tubelets on TRG module and STPro (our approach). }
 	\label{tab:compare_frame_tubelet}
\end{table}

\noindent \textbf{Analysis on temporal curriculum} We examine three TCL strategies based on CNM\cite{cnm}: 1) \textit{Beta Increment} - increasing the distance between positive and negative sample distributions, 2) \textit{Self-Advised} - using model reconstruction scores for easy-to-hard learning, and 3) \textit{Sub-actions} - gradually learning action composition. Table \ref{tab:discuss_tcl_scl} compares tIoU scores, showing \textit{Sub-actions} achieves the highest performance, indicating the model's lack of inherent compositional understanding. This training approach instills compositionality, improving tIoU by 1\%.

\noindent \textbf{Analysis on spatial curriculum} Table \ref{tab:discuss_tcl_scl} shows that both SLF and CGS boosts the performance standalone and jointly when applied in addition with SRM. It shows that both approaches provide orthogonal information which increases model's spatial discriminative ability.

\noindent \textbf{Tubelets vs Frames} STPro, a tubelet-based approach, outperforms frame-based methods for WSTVG. While SRM inference can be done at the frame level, Table \ref{tab:compare_frame_tubelet} shows higher vIoU@0.1 scores relative to vIoU@0.3/vIoU@0.5 for frame-based methods, suggesting they exploit spatial overlap rather than true localization. This confirms tubelets' superiority for spatio-temporal grounding.

\section{Conclusion}
\label{sec:conclude}
This work addresses the challenging task of Weakly Supervised Spatio-Temporal Video Grounding (WSTVG). Leveraging recent advances in vision-language foundation models, we find that simple adaptations fail to provide essential spatio-temporal grounding. To bridge this gap, we introduce Tubelet Referral Grounding (TRG), linking textual queries to tubelets for spatio-temporal predictions. However, TRG struggles with compositional action understanding and complex scenes. To overcome these, we propose STPro, a progressive learning framework with two modules: Sub-Action Temporal Curriculum Learning (SA-TCL) for compositional action understanding, and Congestion-Guided Spatial Curriculum Learning (CG-SCL) for adapting to complex scenes. STPro achieves state-of-the-art performance on three benchmark datasets, improving by 1.0\% on VidSTG-Declarative and 3.0\% on HCSTVG-v1.

{
    \small
    \bibliographystyle{ieeenat_fullname}
    \bibliography{main}
}


\clearpage
\appendix

\setcounter{section}{0}
\setcounter{figure}{0}
\setcounter{table}{0}

\maketitlesupplementary

Here, we provide more details about our approach, additional results, and visual analysis. We also include and expand tables we could not include in the main paper due to space limitations. In summary the supplementary includes the following:

\begin{itemize}

\item Section~\ref{sec:pre_processing_challenges} sheds light on the challenges of using detector and tracker to generate proposal tubelets. We also explore methods to denoise training data and results of the same.
\item Section~\ref{sec:sa_tcl_cg_scl_construction} explores the construction of datasets for SA-TCL \& CG-SCL accompanied by ablations to demonstrate their effectiveness.
\item Section~\ref{sec:vg_2_ablation_and_upper_bound} shows ablation study on HCSTVG-2 dataset.
\item Section~\ref{sec:stpro_vs_fully_sup} compares STPro with fully-supervised approaches across all datasets.
\item Section~\ref{sec:inference_analysis} presents ablations on the filtering criteria used in joint spatio-temporal inference of STPro. We also measure the effects of SLF at inference time and explain the challenges in extending our inference method directly to VidSTG.
\item Section~\ref{sec:slf_in_practice} discusses how Soft-Label Filtering (SLF), a component of CG-SCL is implemented.
\item Section~\ref{sec:pre_proc_prompts} shows the prompts used for the extraction of $\mathtt{POS}$ and sub-action phrases from original captions via an LLM. It also captures some failure cases.
\item Section~\ref{sec:qualitative_analysis} visually demonstrates qualitative improvements brought about by STPro's individual components (SA-TCL \& CG-SCL).
\end{itemize}

\section{Pre-processing Challenges}

\noindent \textbf{Upper-Bound Analysis:} STVG datasets present considerable challenges, with detection and tracking often falling short, as evidenced by the maximum upper bounds in Table~\ref{tab:ab_upper}. Datasets such as HCSTVG and VidSTG feature rapid zooms, scene shifts, occlusions, and defocusing—conditions under which even state-of-the-art detectors struggle to perform reliably. Moreover, the pre-processing step of tracking detections to generate tubelets introduces additional noise. Challenges such as person crossover, abrupt scene transitions (where large bounding box displacements lead to ID mismatches), viewpoint changes, and instances where only part of the body is visible further complicate the task.

Table \ref{tab:ab_upper} presents an upper bound analysis on TRG following two schemes. To the left, we present the upper bound analysis by finding the maximum overlapping detection from Grounding-DINO to the ground-truth bounding box for every frame within the ground-truth temporal boundary. Hence, we consider maximum spatial and temporal overlap. To the right, we present the upper bound analysis by similarly finding which proposed tubelet (as obtained after detection and tracking) has maximum vIoU with the ground-truth tubelet and consider it to be the solution for this sample. There's some decline in performance since tracker adds another level of complexity. We find that our upper bound surpasses the performance of current state-of-the-art fully-supervised approaches by significant margins. 

\begin{table}[htbp]
			\centering
			\renewcommand{\arraystretch}{1.1}
			\scalebox{0.8}{
				\begin{tabular}{c ccc | ccc}
					\specialrule{1.5pt}{0pt}{0pt}
					\rowcolor{mygray} 
					Dataset & m\_tIoU & m\_vIoU &vIoU@0.5 & m\_tIoU & m\_vIoU &vIoU@0.5 \\ \hline\hline
    \cellcolor{mygray} & \multicolumn{3}{c|}{ \cellcolor{mygray} Detection}  & \multicolumn{3}{c}{ \cellcolor{mygray} Post-Tracking}\\ \hline
				HCSTVG-v1 & 92.1 & 69.3 & 86.2 & 83.6 & 65.0 & 76.3\\
                    HCSTVG-v2 & 95.1 & 68.8 & 87.0 & 81.5 & 60.0 & 68.0\\
                     VidSTG-D & 83.8 & 52.7 & 60.9 & 72.5 & 45.4 & 50.6\\
                     VidSTG-I & 83.7 & 48.8 & 54.7 & 72.7 & 41.6 & 44.6\\ 
				\specialrule{1.5pt}{0pt}{0pt}
			\end{tabular}}
   \caption{\textbf{Upper-bound analysis} on HCSTVG-1, HCSTVG-2, and VidSTG utilizing only per-frame detections (left) and tubelets (right) as proposals.}
   \label{tab:ab_upper}
\end{table}

\label{sec:pre_processing_challenges}

\noindent \textbf{Tubelets Pre-processing} 
In this section, we detail the pre-processing steps to clean the training data used in SRM for learning referral subject grounding. A significant issue identified during this process is the prevalence of soft-label switching within tubelets. Specifically, a tubelet often contains multiple detections, each with a different soft-label generated by Grounding-DINO. 

\begin{figure}[htbp]
    \centering
    \includegraphics[width=\linewidth]{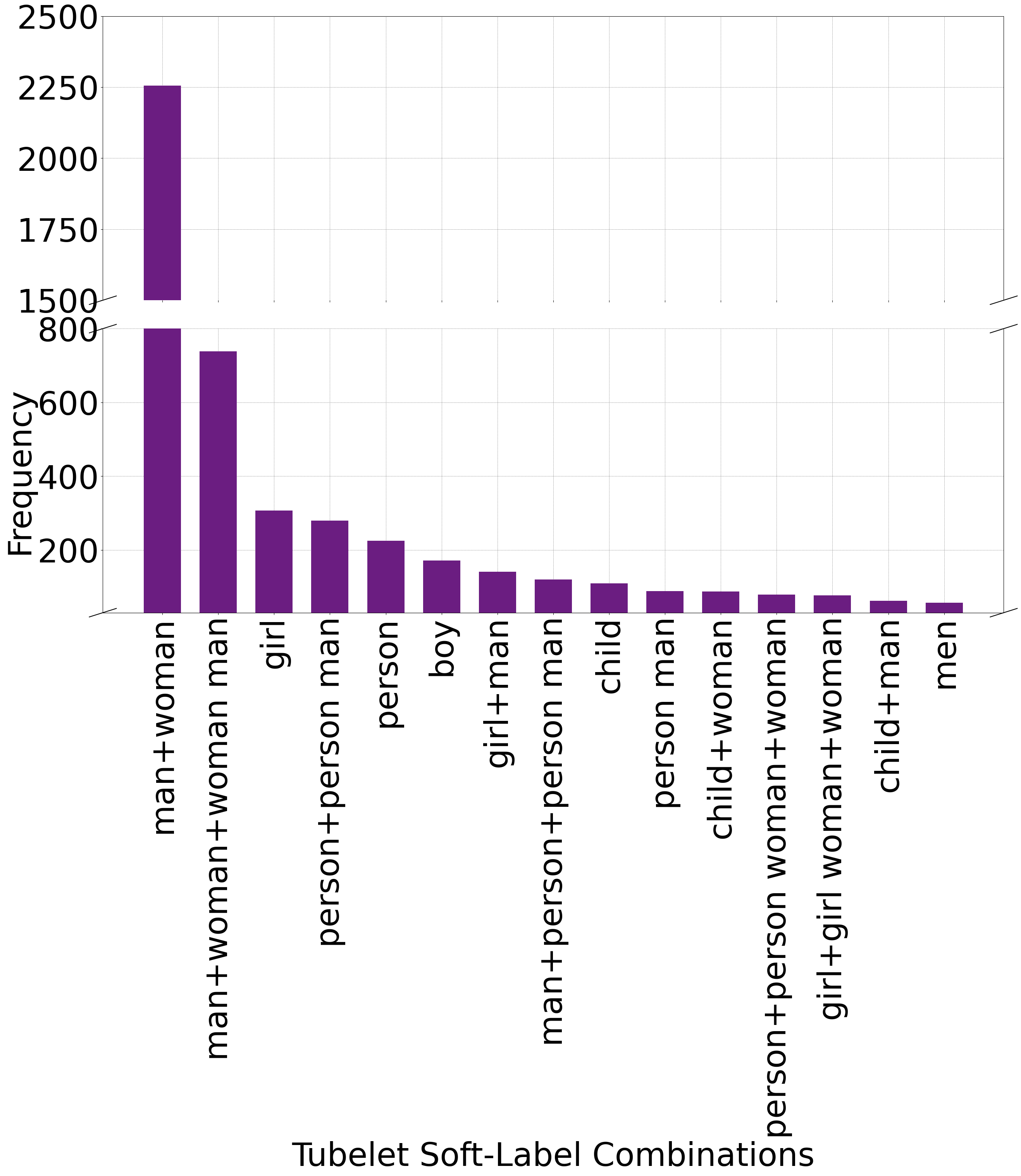}
    \caption{Distribution of frequently co-occurring soft-labels in tubelets for HCSTVG-1. Grounding-DINO soft labels can be more than one class for a given detection (e.g., person man).}
    \label{fig:soft_label_combinations}
\end{figure}

Figure~\ref{fig:soft_label_combinations} illustrates some common combinations of soft-labels observed in tubelets. Examples like \textit{man+woman} and \textit{girl+man} indicate that some tubelets do not consistently track a single individual but instead switch subjects temporally. 
Such label inconsistencies can degrade training performance, especially when subjects belong to different categories. For instance, if a tubelet switches between \textit{man} and \textit{woman}, the model learns to ground both categories simultaneously, despite the referred subject being explicitly a \textit{man}. This creates a noisy training signal, undermining the model's ability to learn accurate grounding.

To address this, we first quantify the extent of soft-label switching in Section~\ref{sec:switching_percentage}. Subsequently, in Section~\ref{sec:switching_iou_distribution}, we propose leveraging soft-labels in conjunction with Intersection over Union (IoU) metrics to detect cases where subject switching actually occurs within tubelets. In Section~\ref{sec:switching_iou_distribution} we analyze the temporal duration over which such switches take place within a parent tubelet. Finally in Section~\ref{sec:training_set_denoising_strategies} (based on analysis in the previous two sections) we devise and employ dataset denoising strategies aimed at improving training data quality and, consequently, model grounding capabilities.

\subsection{Percentage Switching in Train} 
\label{sec:switching_percentage}
\begin{figure}[htbp]
    \centering
    \includegraphics[width=0.9\linewidth]{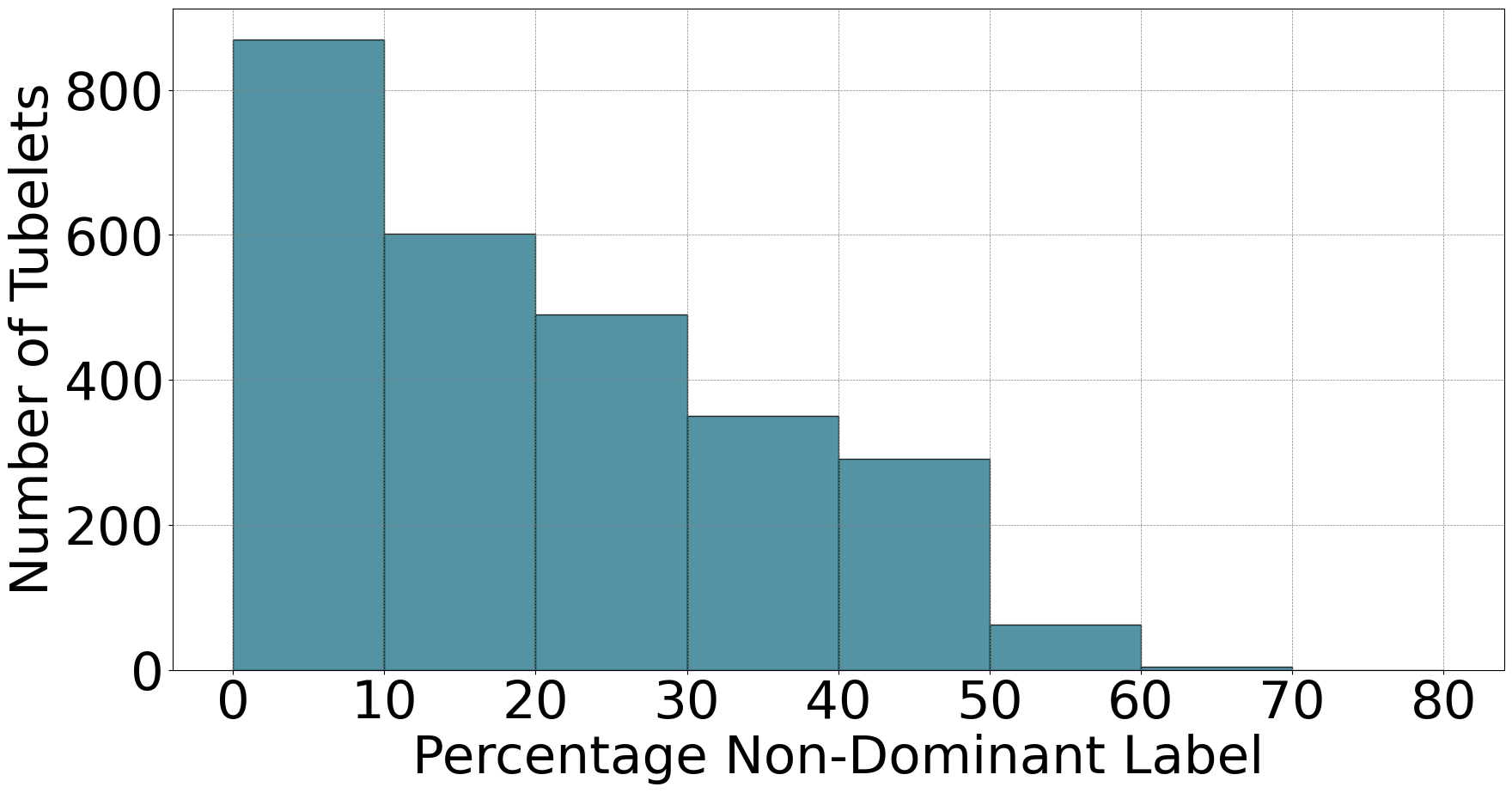}
    \caption{Distribution of the percentage of non-dominant soft-label detections in tubelets for all tubelets with conflicting combinations (e.g., man + woman) in HCSTVG-1.}
    \label{fig:percentage_switching_in_faulty_tubelets}
\end{figure}
In Figure~\ref{fig:soft_label_combinations}, we observe several instances where soft-label switching occurs within tubelets. For tubelets with conflicting combinations (e.g., \textit{man+woman}), Figure~\ref{fig:percentage_switching_in_faulty_tubelets} illustrates the percentage of detections in each tubelet that do not correspond to the most frequently occurring soft-label. Our analysis reveals that while the majority of tubelets with conflicting combinations exhibit less than 10\% switching, a significant portion contains more than 30\% switching. This highlights the presence of considerable noise in the training dataset, underscoring the need for data cleaning to ensure a high-quality training signal for the model.
 
\subsection{Label Switched Sub-Tubelet Analysis} 
\label{sec:switching_iou_distribution}
\begin{figure}[t!]
    \centering
    \includegraphics[width=0.9\linewidth]{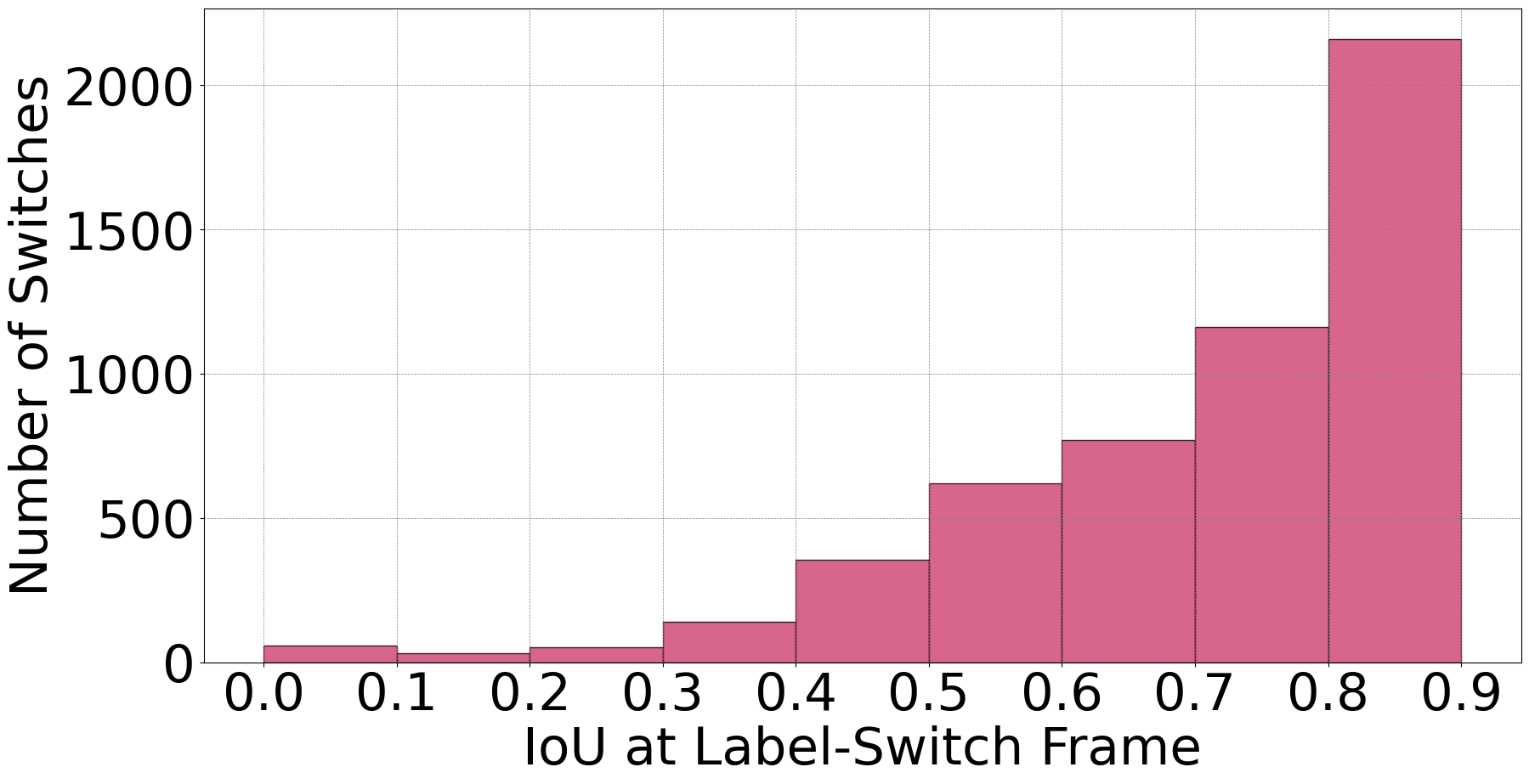}
    \caption{Distribution of the IoU between detections on the frame before and after faulty soft-label switch in HCSTVG-1.}
    \label{fig:switching_per_frame_percentages}
\end{figure}
Fig~\ref{fig:percentage_switching_in_faulty_tubelets} highlights the extent of switching in tubelets with conflicting label combinations. We hypothesize that in cases where the tracked subject switches, the IoU overlap between the detection of the subject in the frame preceding the label switch and the detection at the switching frame should be low. To investigate whether IoU can reliably identify instances of visual subject-switching as opposed to faulty soft-label switches, we analyze the distribution of IoU values at these switching points for all tubelets with conflicting label combinations. This distribution is presented in Figure~\ref{fig:switching_per_frame_percentages}.

The results indicate that a substantial number of switching points exhibit IoU values below 0.50, suggesting that the tracked subject may have visually switched, rather than the switch being solely a result of faulty soft-labeling. However, a significant number of high IoU values at switch points indicate that Grounding-DINO may sometimes produce faulty label switches when subjects overlap, even while the original subject continues to be correctly tracked. Differentiating between faulty soft-label switches and visual-subject switches in the case of overlapping subjects, however, cannot reliably be performed using this technique.

\begin{figure}[H]
    \centering
    \includegraphics[width=0.9\linewidth]{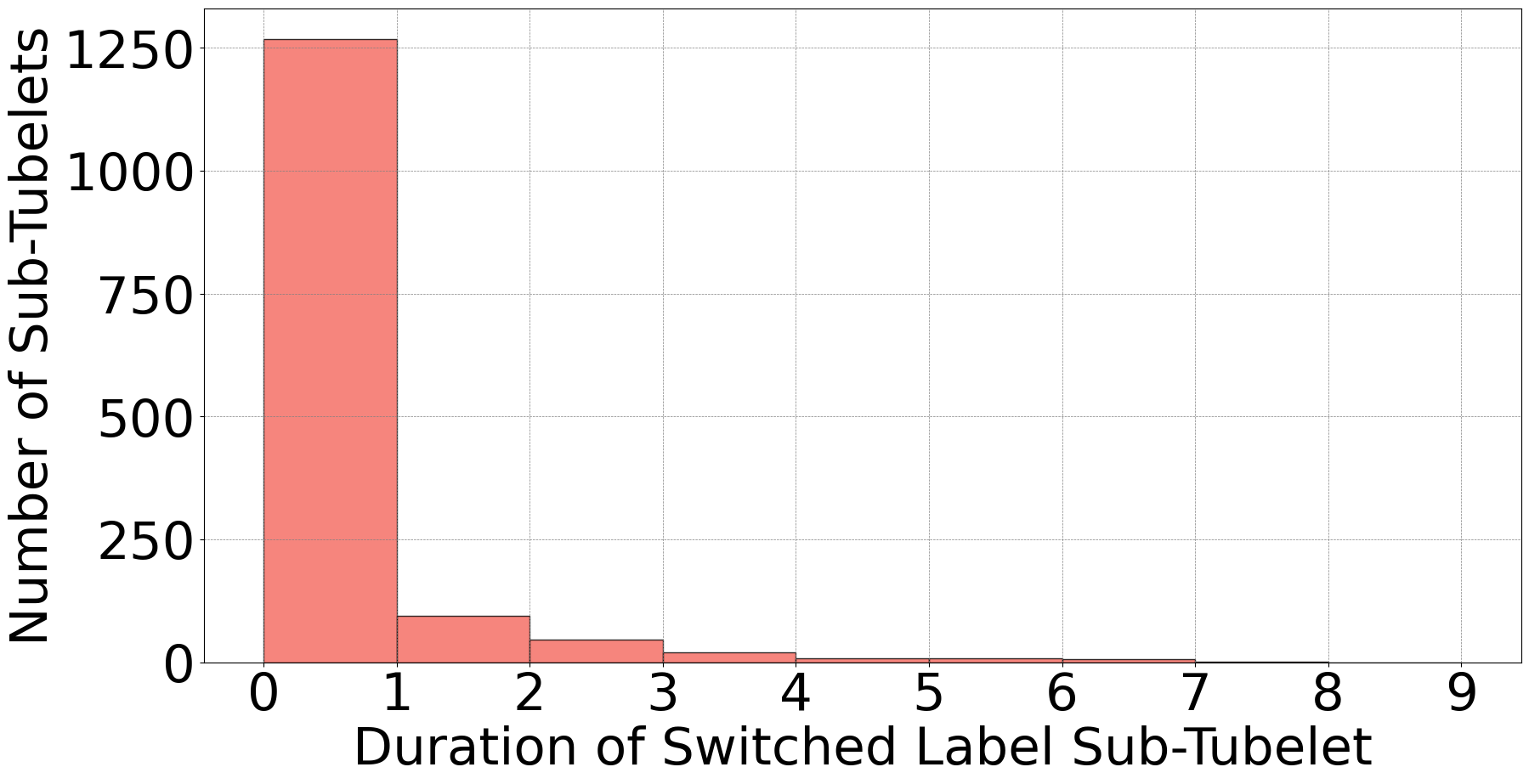}
    \caption{Distribution of durations (in seconds) for soft-label switched sub-tubelets for tubelets with conflicting combinations in HCSTVG-1. For each tubelet we find continuous sub-sections within the tubelet where soft-label switch occurs and find its duration.}
    \label{fig:switched_tubelet_durations}
\end{figure}

Our analysis reveals that when a label switch occurs, the sub-tubelet associated with the new soft label is typically less than 1 second in duration, as shown in Figure~\ref{fig:switched_tubelet_durations}. This indicates that label switches often manifest as brief, intermittent segments within the parent tubelet, rather than representing a permanent change in the tracked subject. Such transient switches highlight the need for targeted strategies to address these short-lived inconsistencies.

\subsection{Training Set Denoising Strategies}
\label{sec:training_set_denoising_strategies}
To address the primary challenges, we propose two de-noising strategies based on the heuristics from Section~\ref{sec:switching_iou_distribution}.

\noindent \textbf{Switch-Addition:} This strategy focuses on sub-tubelets with durations exceeding 1 second. Sub-tubelets meeting this threshold are extracted and converted into independent tubelets, while the remaining non-mode labeled detections are eliminated. This allows sub-tubelets to be grounded independently. A consequence of this choice is that some samples have an increased number of tubelets to distinguish over possibly making training more challenging.

\noindent \textbf{Switch-Dropping:} This strategy eliminates all detections within a tubelet (conflicting combination tubelets only) that do not correspond to the mode label. The goal of this strategy is to remove maximum cases in which visual-switching occurs. However this strategy may result in the elimination of certain continuous stretches of switched label (termed sub-tubelet) which correspond to the optimal tubelet to be grounded for that sample.

We employe both of these de-noising strategies in our TRG training pipeline and summarize the results in Table~\ref{tab:denoising_strategies}. Both strategies improve the performance over the baseline. Switch-dropping outperforms switch-addition strategy. 

\begin{table}[H]
    \centering
    \renewcommand{\arraystretch}{1.1}
    \scalebox{0.9}{
        \begin{tabular}{c | cccc}
            \specialrule{1.5pt}{0pt}{0pt}
            \rowcolor{mygray} 
   \rowcolor{mygray} Method & m\_vIoU & vIoU@0.1 & vIoU@0.3 & vIoU@0.5 \\ \hline\hline
    Baseline  & 10.40 & 30.69 & 12.67 & 5.00 \\
    Switch-Addition & 9.74 & 29.31 & 11.81 & 4.40 \\
    Switch-Dropping & \textbf{10.81} & 
    \textbf{31.47} & \textbf{13.19} & \textbf{5.43} \\
     \specialrule{1.5pt}{0pt}{0pt}
\end{tabular}}
\caption{Comparison and analysis of train set denoising strategies for TRG on HCSTVG-1.}
\label{tab:denoising_strategies}
\end{table}

\section{SA-TCL \& CG-SCL Stage-wise Construction}
\label{sec:sa_tcl_cg_scl_construction}

In Section~\ref{sec:cgs_interval_selection}, we first analyze SRM model performance based on different configurations of pairwise average temporal IoU (between tubelets) to find the ideal hyper-parameters for stage-wise CGS construction. In Section \ref{sec:extracted_actions_for_tcl} we ponder over our temporal curriculum (SA-TCL) to better understand why training benefits from individual actions to action-combinations and not the opposite. Finally, we explore whether it is beneficial to have cumulative training stages (each stage comprises all training samples from preceding stages) for CG-SCL and SA-TCL in Section \ref{sec:overlapping_substages}.
The resulting stage-wise dataset statistics based on the below analysis can be found in Figure~\ref{fig:dataset_stats_r1} and Figure~\ref{fig:dataset_stats_r2}.

\subsection{Interval Selection for CGS} 
\label{sec:cgs_interval_selection}
\begin{table}[htbp]
	\centering
\renewcommand{\arraystretch}{1.1}
\scalebox{0.7}{
\begin{tabular}{cc | cccc}
    \rowcolor{mygray} 
    \specialrule{1.5pt}{0pt}{0pt}
    Stages & IoU/Stage & m\_tIoU & m\_vIoU & vIoU@0.3 &  vIoU@0.5  \\ 
    \hline\hline
     \multicolumn{6}{c}{\textit{Low to High Temporal IoU overlap}}  \\ \hline 
        7 & 0.14 & 17.63 & \textbf{9.30} & \textbf{12.16} & \textbf{4.14} \\
        5 & 0.20 & \textbf{18.89} & 9.24 & 11.64 & \textbf{4.14} \\
        3 & 0.33 & 18.51 & 9.08 & 11.47 & 4.31 \\
    \hline
    \multicolumn{6}{c}{\textit{High to Low Temporal IoU overlap}} \\ \hline 
        7 & 0.14 & \textbf{21.88} & \textbf{11.10} & \textbf{14.05} & \textbf{5.60}  \\
        5 & 0.20 & 21.13 & 11.01 & 13.36 & 5.43 \\
        3 & 0.33 & 20.14 & 10.36 & 12.67 & 5.00 \\
 \specialrule{1.5pt}{0pt}{0pt}
\end{tabular}}
  \caption{Ablation on number of curriculum stages at different temporal IoU overlap/stage in both low-to-high and high-to-low settings for HCSTVG-1.}
    \label{tab:cgs_thresholding_summary}
\end{table}

\begin{figure}[H]
    \centering
    \includegraphics[width=0.9\linewidth]{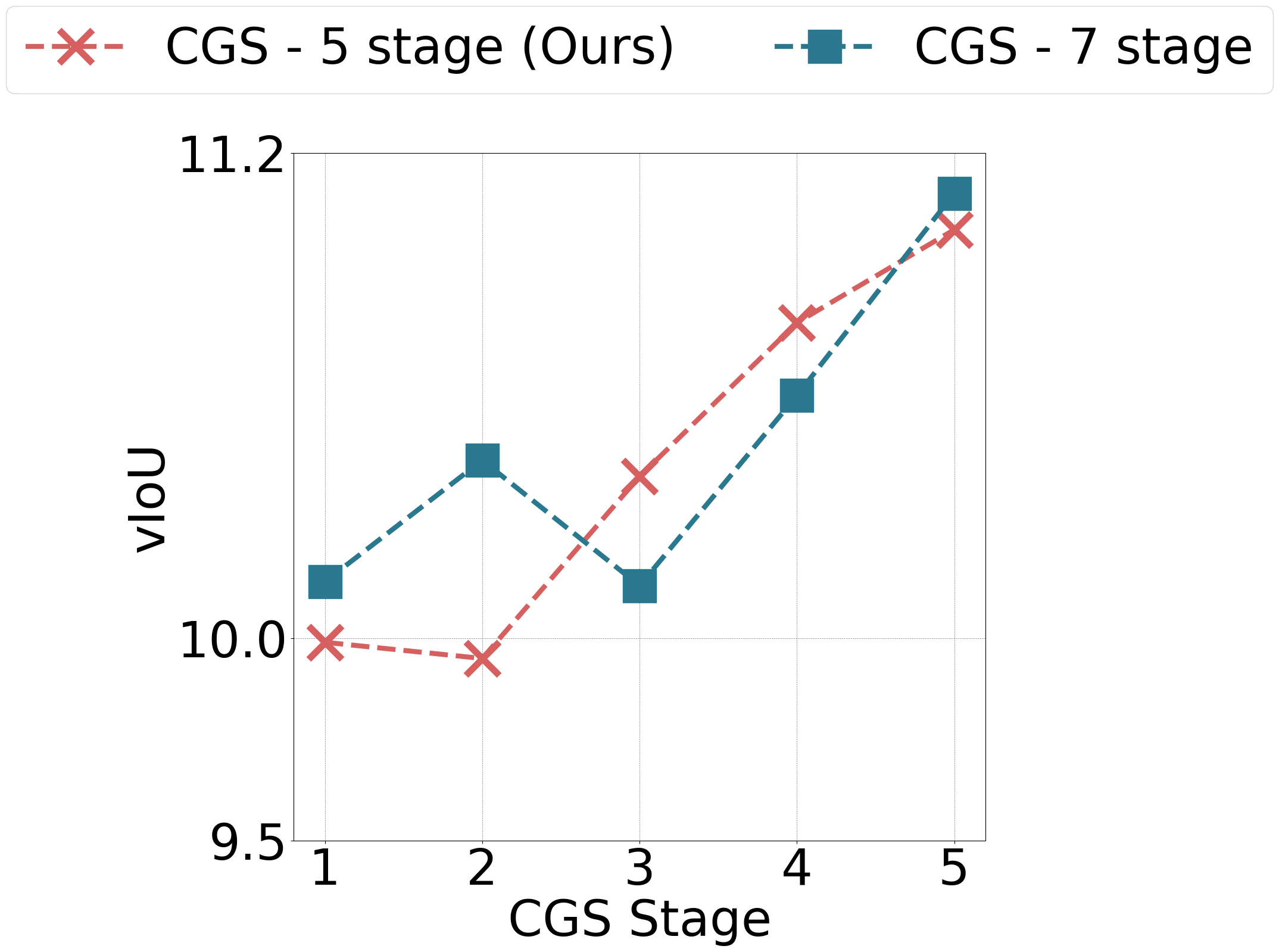}
    \caption{Per-stage performance gain for 5-stage and 7-stage (high-to-low) CGS for HCSTVG-1.}
    \label{fig:five_stage_vs_seven_stage}
\end{figure}
Intuitively, using temporally spaced candidate tubelets aids in learning referral subject localization by making it easier to distinguish attributes and actor types due to their diversity in features. Conversely, less diverse candidates might help the model focus on actor differentiation while minimizing background variation, promoting the gradual divergence of unrelated features in the joint semantic space. To evaluate these hypotheses, we conduct a systematic study of two settings. We tested two strategies: gradually increasing the per-stage temporal IoU threshold from low-to-high and high-to-low (Ours CGS method). From Table ~\ref{tab:cgs_thresholding_summary}, we observe that progressing from high-to-low temporal IoU improves the model's referral capabilities.
We also observe that using 5 stages with a per-stage delta of 0.20 IoU results in consistent improvements at each stage, as expected by the curriculum. Although using 7 stages increases the model's overall capability, the per-stage progress is less consistent (see Figure~\ref{fig:five_stage_vs_seven_stage}), and the training time increases substantially.

\begin{figure}[H]
    \centering
    \includegraphics[width=0.9\linewidth]{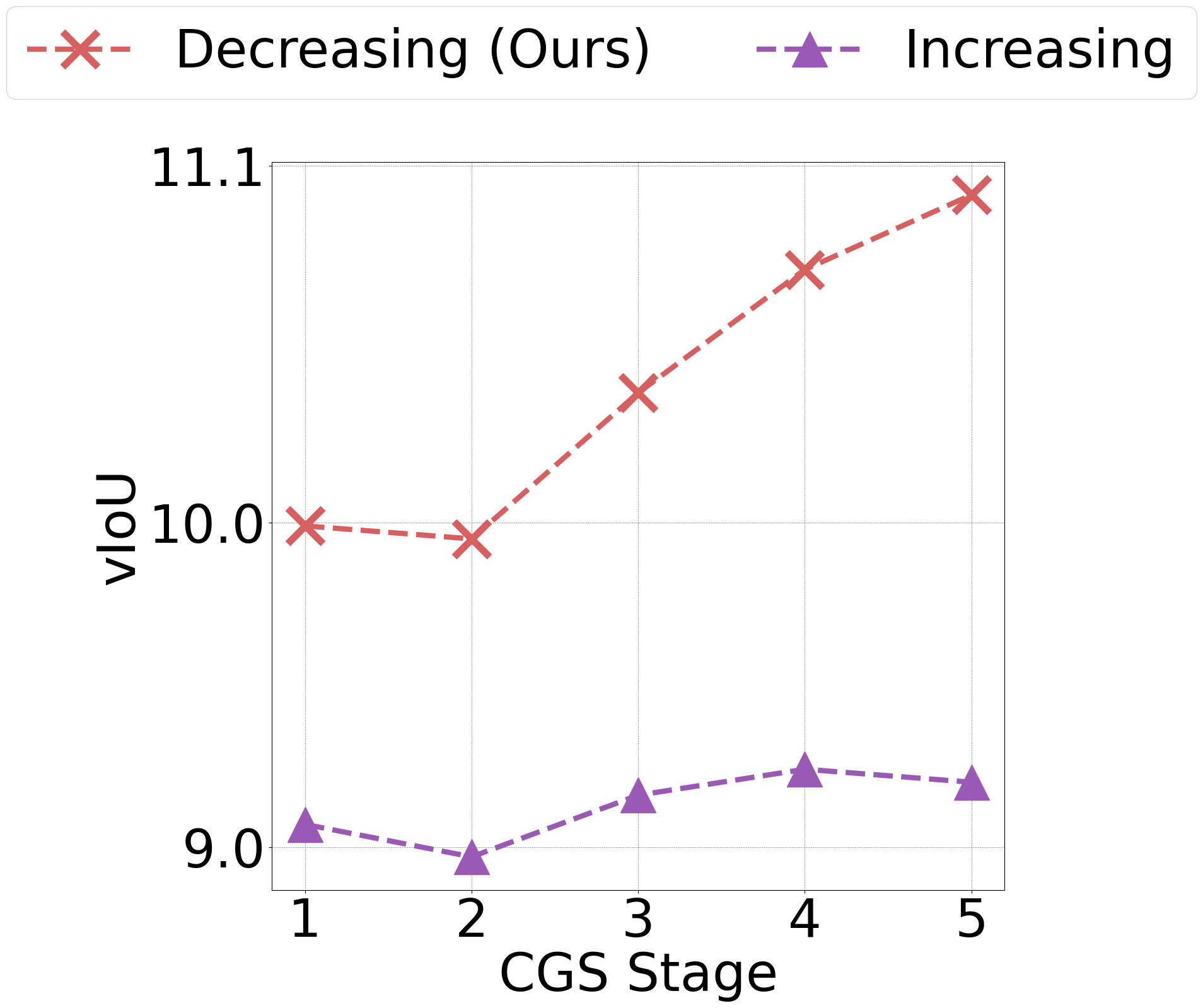}
    \caption{Per-stage SRM performance improvement in 5-stage low-to-high and high-to-low CGS for HCSTVG-1. In the low-to-high CGS, each stage yields minor increments}
    \label{fig:cgs_inc_vs_dec}
\end{figure}
Furthermore, while progressing from low to high IoU shows some relative improvement, Figure~\ref{fig:cgs_inc_vs_dec} highlights a notable difference: decreasing the IoU (high-to-low) leads to a consistent improvement in model capabilities at each stage. This pattern is absent when increasing IoU (low-to-high), suggesting that the model progressively learns to differentiate actors in similar background contexts before handling more diverse contexts.

\subsection{Utility of Extracted Actions for SA-TCL}
\label{sec:extracted_actions_for_tcl}
We first examine the benefits of deconstructing complex captions and using their sub-components for training in SA-TCL. Specifically, we compare the temporal localization performance of TRM using SA-TCL with a curriculum where we progressively increase the number of actions in the original captions without extracting sub-action phrases. As shown in Table \ref{tab:original_captions_vs_extracted_captions}
, the model benefits from the extracted sub-captions, gaining compositional capabilities that are not learned by simply increasing the number of actions.

\begin{table}[H]
    \centering
    \renewcommand{\arraystretch}{1.1}
    \scalebox{0.8}{
        \begin{tabular}{c | cccc}
            \specialrule{1.5pt}{0pt}{0pt}
            \rowcolor{mygray} 
   \rowcolor{mygray} Method & m\_tIoU & tIoU@0.1 & tIoU@0.3 & tIoU@0.5 \\ \hline\hline
    Baseline  & 32.99 & 78.43 & 56.43 & 22.52 \\
    Original Captions & 33.05 & 77.48 & 57.03 & 22.78 \\
    Extracted Captions & \textbf{33.92} & \textbf{78.77} & \textbf{57.72} & \textbf{24.68} \\
     \specialrule{1.5pt}{0pt}{0pt}
\end{tabular}}
\caption{Ablation on the use of extracted sub-actions compared to original captions for increasing action SA-TCL (HCSTVG-1).}
\label{tab:original_captions_vs_extracted_captions}
\end{table}
Next, we investigate whether construction or deconstruction is more effective for learning action composition in SA-TCL. We reverse the order of sub-action phrases during training, starting with compound phrases and gradually adding individual actions (Dec in Table \ref{tab:sa-tcl_analysis}). In contrast, the original SA-TCL method (Inc in Table~\ref{tab:sa-tcl_analysis}) increases action composition progressively. Our results show that construction (Inc) significantly outperforms deconstruction (Dec) in understanding action compositionality. We hypothesize that this is due to the model's difficulty in reducing its predicted temporal span in later training stages.

\begin{table}[H]
    \centering
    \renewcommand{\arraystretch}{1.1}
    \scalebox{0.8}{
        \begin{tabular}{c | c c c c}
            \specialrule{1.5pt}{0pt}{0pt}
            \rowcolor{mygray} 
            Curriculum & m\_tIoU & tIoU@0.1 & tIoU@0.3 & tIoU@0.5 \\ \hline\hline
            Dec & 31.41 & 74.03 & 53.15 & 22.69 \\
            Baseline & 32.99 & 78.43 & 56.43 & 22.52 \\
            Inc & \textbf{33.92} & \textbf{78.77} & \textbf{57.72} & \textbf{24.68} \\
            \specialrule{1.5pt}{0pt}{0pt}
        \end{tabular}}
    \caption{Comparing re-construction (Inc) and de-construction (Dec) using extracted sub-actions for SA-TCL (HCSTVG-1).}
    \label{tab:sa-tcl_analysis}
\end{table}

\subsection{Merit of Overlapping Sub-stages}
\label{sec:overlapping_substages}
Table \ref{tab:tcl_and_cgs_overlap_comparison} compares overlapping substages in SA-TCL and CGS (in CG-SCL), with the first row showing results for TRM and SRM without any curriculum. Our analysis reveals that overlapping substages, by incorporating training samples from previous stages (stages already trained over), helps prevent forgetting and offer benefits. 

\begin{table}[h]
    \centering
    \renewcommand{\arraystretch}{1.1}
    \scalebox{0.77}{
        \begin{tabular}{c | c | c c c c}
            \specialrule{1.5pt}{0pt}{0pt}
            \rowcolor{mygray} 
            Curriculum & Cumulative & m\_tIoU & tIoU@0.1 & tIoU@0.3 & tIoU@0.5 \\ \hline\hline
            SA-TCL (Inc) & - & 32.99 & 78.43 & 56.43 & 22.52 \\
            SA-TCL (Inc) & & 33.74 & 80.41 & 57.64 & 23.81 \\
            SA-TCL (Inc) & \checkmark & \textbf{33.92} & \textbf{78.77} & \textbf{57.72} & \textbf{24.68} \\
            \specialrule{1.5pt}{0pt}{0pt}
            \rowcolor{mygray} 
            Curriculum & Cumulative & m\_vIoU & vIoU@0.1 & vIoU@0.3 & vIoU@0.5 \\ \hline\hline
            CGS (Dec) & - & 10.40 & 30.69 & 12.67 & 5.00 \\
            CGS (Dec) & & 10.79 & 31.29 & 13.19 & 5.43 \\
            CGS (Dec) & \checkmark & \textbf{11.01} & \textbf{32.41} & \textbf{13.36} & \textbf{5.43} \\
            \specialrule{1.5pt}{0pt}{0pt}
        \end{tabular}}
    \caption{Comparing the use of overlapping (cummulative) sub-stages in SA-TCL \& CG-SCL for HCSTVG-1. The first row (in upper and lower) represents the baseline scores from TRM and SRM.}
    \label{tab:tcl_and_cgs_overlap_comparison}
\end{table}

\begin{figure*}
    \centering
    \renewcommand{\arraystretch}{1.1}
    \scalebox{0.77}{
        \begin{subtable}[t]{0.30\textwidth}
            \centering
            \begin{tabular}{c | cc}
                \specialrule{1.5pt}{0pt}{0pt}
                \rowcolor{gray!20} Stage & Additional & Total \\ \hline\hline
                Stage-1 & 356 & 356 \\
                Stage-2 & 140 & 496 \\
                Stage-3 & 292 & 788 \\
                Stage-4 & 1560 & 2348 \\
                Stage-5 & 2069 & 4417 \\
                \specialrule{1.5pt}{0pt}{0pt}
            \end{tabular}
            \caption{\textbf{HCSTVG-1:} CGS stage-wise}
            \label{tab:vg1_cgs}
        \end{subtable}%
        \hspace{0.075cm}
        \begin{subtable}[t]{0.30\textwidth}
            \centering
            \begin{tabular}{c | cc}
                \specialrule{1.5pt}{0pt}{0pt}
                \rowcolor{gray!20} Stage & Additional & Total \\ \hline\hline
                Stage-1 & 302 & 302 \\
                Stage-2 & 125 & 427 \\
                Stage-3 & 255 & 682 \\
                Stage-4 & 1354 & 2036 \\
                Stage-5 & 2129 & 4165 \\
                \specialrule{1.5pt}{0pt}{0pt}
            \end{tabular}
            \caption{\textbf{HCSTVG-1:} CGS + SLF stage-wise}
            \label{tab:vg1_cgs_slf}
        \end{subtable}%
        \hspace{0.075cm}

        \begin{subtable}[t]{0.30\textwidth}
            \centering
            \begin{tabular}{c | cc}
                \specialrule{1.5pt}{0pt}{0pt}
                \rowcolor{gray!20} Stage & Additional & Total \\ \hline\hline
                Stage-1 & 882 & 882 \\
                Stage-2 & 318 & 1200 \\
                Stage-3 & 707 & 1907 \\
                Stage-4 & 3548 & 5491 \\
                Stage-5 & 4604 & 10095 \\
                \specialrule{1.5pt}{0pt}{0pt}
            \end{tabular}
            \caption{\textbf{HCSTVG-2:} CGS stage-wise}
            \label{tab:vg2_cgs}
        \end{subtable}%
        \hspace{0.075cm}
        \begin{subtable}[t]{0.30\textwidth}
            \centering
            \begin{tabular}{c | cc}
                \specialrule{1.5pt}{0pt}{0pt}
                \rowcolor{gray!20} Stage & Additional & Total \\ \hline\hline
                Stage-1 & 721 & 721 \\
                Stage-2 & 288 & 1009 \\
                Stage-3 & 617 & 1626 \\
                Stage-4 & 3057 & 4683 \\
                Stage-5 & 4825 & 9508 \\
                \specialrule{1.5pt}{0pt}{0pt}
            \end{tabular}
            \caption{\textbf{HCSTVG-2:} CGS+SLF stage-wise}
            \label{tab:vg2_cgs_slf}
        \end{subtable}%
    }
    \caption{CG-SCL per-stage and cumulative dataset statistics for HCSTVG-1 \& HCSTVG-2.}
    \label{fig:dataset_stats_r1}
\end{figure*}
\begin{figure*}
    \centering
    \renewcommand{\arraystretch}{1.1}
    \scalebox{0.77}{
        \begin{subtable}[t]{0.30\textwidth}
            \centering
            \begin{tabular}{c | cc}
                \specialrule{1.5pt}{0pt}{0pt}
                \rowcolor{gray!20} Stage & Additional & Total \\ \hline\hline
                Stage-1 & 974 & 974 \\
                Stage-2 & 2068 & 3042 \\
                Stage-3 & 1251 & 4293 \\
                Stage-4 & 182 & 4475 \\
                \specialrule{1.5pt}{0pt}{0pt}
            \end{tabular}
            \caption{\textbf{HCSTVG-1:} Original captions}
            \label{tab:vg1_tcl_orig}
        \end{subtable}%
        \hspace{0.075cm}
        \begin{subtable}[t]{0.30\textwidth}
            \centering
            \begin{tabular}{c | cc}
                \specialrule{1.5pt}{0pt}{0pt}
                \rowcolor{gray!20} Stage & Additional & Total \\ \hline\hline
                Stage-1 & 11394 & 11394 \\
                Stage-2 & 7045 & 18439 \\
                Stage-3 & 2903 & 21342 \\
                Stage-4 & 952 & 22294 \\
                \specialrule{1.5pt}{0pt}{0pt}
            \end{tabular}
            \caption{\textbf{HCSTVG-1:} Extracted captions}
            \label{tab:vg1_tcl_extracted}
        \end{subtable}%
        \hspace{0.075cm}

        \begin{subtable}[t]{0.30\textwidth}
            \centering
            \begin{tabular}{c | cc}
                \specialrule{1.5pt}{0pt}{0pt}
                \rowcolor{gray!20} Stage & Additional & Total \\ \hline\hline
                Stage-1 & 806 & 806 \\
                Stage-2 & 4532 & 5338 \\
                Stage-3 & 3270 & 8608 \\
                Stage-4 & 1386 & 9994 \\
                \specialrule{1.5pt}{0pt}{0pt}
            \end{tabular}
            \caption{\textbf{HCSTVG-2:} Original captions}
            \label{tab:vg2_tcl_orig}
        \end{subtable}%
        \hspace{0.075cm}
        \begin{subtable}[t]{0.30\textwidth}
            \centering
            \begin{tabular}{c | cc}
                \specialrule{1.5pt}{0pt}{0pt}
                \rowcolor{gray!20} Stage & Additional & Total \\ \hline\hline
                Stage-1 & 26211 & 26211 \\
                Stage-2 & 18042 & 44253 \\
                Stage-3 & 6846 & 51099 \\
                Stage-4 & 2253 & 53352 \\
                \specialrule{1.5pt}{0pt}{0pt}
            \end{tabular}
            \caption{\textbf{HCSTVG-2:} Extracted captions}
            \label{tab:vg2_tcl_extracted}
        \end{subtable}%
    }
    \caption{SA-TCL per-stage and cummulative statistics for HCSTVG-1 \& HCSTVG-2.}
    \label{fig:dataset_stats_r2}
\end{figure*}

\section{Ablation study on HCSTVG-v2 dataset}

\label{sec:vg_2_ablation_and_upper_bound}

We show an ablation study of the breakdown of the different components of TRG as well as STPro (SA-TCL \& CG-SCL) on the HCSTVG-v2 dataset, an extended version of HCSTVG-v1 in Table \ref{tab:abla_all_hcstvgv2}. 

\noindent \textbf{Effectiveness of TRG sub-modules} Similar to HCSTVG-v1 dataset, we observe SRM, TRM and $\mathtt{POS}$ improves the performance over weakly adapted Grounding DINO. Combining all sub-modules, the performance boost is 16.35\%, 7.71\%, 14.00\%, and, 6.37\% at mean tIoU, vIoU, vIoU@0.3, and, vIoU@0.5 respectively.

\noindent \textbf{Impact of SA-TCL and CG-SCL} Both TA-SCL and CG-SCL improves TRG's performance independently. On combining both, we find that we do not achieve improvement over CG-SCL alone. However, the performance difference is very minimal (< 0.2\% at mean vIoU). STPro outperforms W-GDINO by a margin of 10.14\% and TRG by a margin of 2.43\% at mean vIoU. 

\begin{figure}[htbp]
\centering
\includegraphics[width=\linewidth]{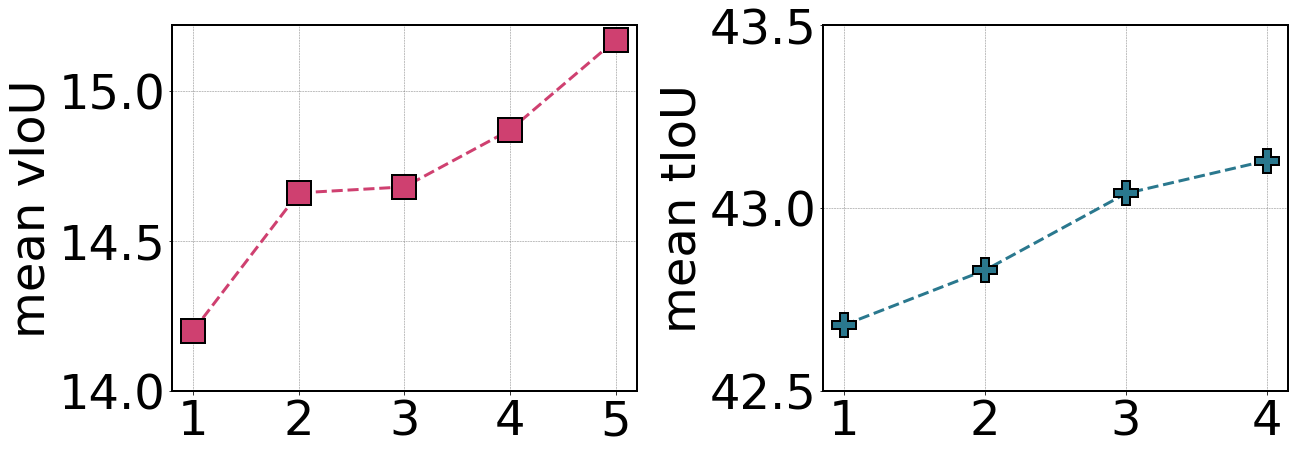}
\caption{Performance at successive stages of \textcolor{myteal}{SA-TCL} and \textcolor{mypink}{CG-SCL} on HCSTVG-v2 dataset.}
\label{fig:clm_stagewise_hcstvgv2}

\end{figure}

\noindent \textbf{Analysis on stages of Curriculum Learning} Fig. \ref{fig:clm_stagewise_hcstvgv2} shows 
increment of score for successive stages of SA-TCL and CG-SCL.

\begin{table}[htbp]
	\centering
\renewcommand{\arraystretch}{1.1}
\scalebox{0.9}{
\begin{tabular}{ccc cccc}
    \rowcolor{mygray} 
    \specialrule{1.5pt}{0pt}{0pt}
    SRM & TRM  & $\texttt{\textbf{POS}}$ & m\_tIoU & m\_vIoU & vIoU@0.3 &  vIoU@0.5  \\ 
    \hline\hline
          &&& 23.30 &  9.85 & 13.30 & 5.63 \\
        \checkmark & & & 30.99 & 10.44 & 13.90 & 6.35 \\
        \checkmark & & \checkmark & 28.97 & 13.62 & 19.15 & 9.00 \\
        \checkmark & \checkmark  & \checkmark & \textbf{39.55} & \textbf{17.56} & \textbf{27.20} & \textbf{12.00} \\
    \specialrule{1.5pt}{0pt}{0pt}
    TRG & CG-SCL & SA-TCL  & m\_tIoU & m\_vIoU & vIoU@0.3 &  vIoU@0.5  \\ 
    \hline\hline 
    &&& 23.30 &  9.85 & 13.30 & 5.63 \\
\checkmark  && & \textbf{39.55} & 17.56 & 27.20 & 12.00 \\
\checkmark & & \checkmark & 39.02 & 17.62 & 27.50 & 12.10 \\
\checkmark & \checkmark && 39.00 & \textbf{19.99} & \textbf{31.70} & \textbf{14.55} \\
\checkmark & \checkmark & \checkmark & 38.46 & 19.82 & 31.35 & 14.25 \\
 \specialrule{1.5pt}{0pt}{0pt}
\end{tabular}}
  \caption{\textbf{Ablation study} on sub-modules of TRG module(upper) and different components of STPro (lower) on HCSTVG-v2 dataset.}
    \label{tab:abla_all_hcstvgv2}

\end{table}

\begin{table*}[h]
	\centering
	\renewcommand{\arraystretch}{1.06}
	\scalebox{0.8}{
		\begin{tabular}{r cccc cccc}
				\rowcolor{mygray} 
				\specialrule{1.5pt}{0pt}{0pt}
    \cellcolor{mygray}  & \multicolumn{4}{c}{ \cellcolor{mygray} HCSTVG - v1} & \multicolumn{4}{c}{ \cellcolor{mygray} HCSTVG - v2} \\ 

                    \rowcolor{mygray} Methods & m\_tIoU & m\_vIoU & vIoU@0.3 & vIoU@0.5  & m\_tIoU & m\_vIoU & vIoU@0.3 &  vIoU@0.5  \\ 
				\hline\hline
    \textit{Fully-Supervised} \\ \hline
    STGVT \textcolor{lightgray}{\scriptsize{[TCSVT20]}}~\citep{hcstvg} & - & 18.2 & 26.8 & 9.5  & - & - & - & - \\
    STVGBert \textcolor{lightgray}{\scriptsize{[ICCV21]}}~\citep{stvgbert} & - & 20.4 & 29.4 &  11.3  & - & - & - & -\\
    TubeDETR \textcolor{lightgray}{\scriptsize{[CVPR22]}}~\citep{Yang2022TubeDETRSV} & 43.7 & 32.4 & 49.8 & 23.5 & 53.9 & 36.4 & 58.8 & 30.6\\
    STCAT \textcolor{lightgray}{\scriptsize{[NeurIPS22]}}~\citep{Jin2022EmbracingCA} & 49.4 & 35.1 & 57.7 & 30.1 & - & - & - & - \\
    CSDVL \textcolor{lightgray}{\scriptsize{[CVPR23]}}~\citep{clb} & - & 36.9 & 62.2 & 34.8 & 58.1 & 38.7 & 65.5 & 33.8 \\ 
            CG-STVG \textcolor{lightgray}{\scriptsize{[CVPR24]}}~\citep{cg-stvg} & 52.8 & 38.4 & 61.5 & 36.3 & 60.0 & 39.5 & 64.5 & 36.3\\
            VGDINO \textcolor{lightgray}{\scriptsize{[CVPR24]}}~\citep{video-gdino} & - & 38.3 & 62.5 & 36.1 & - & 39.9 & 67.1 & 34.5\\ \hline
    \textit{Weakly-Supervised (Two-stage pipelines)} \\ \hline
     GroundeR \textcolor{lightgray}{\scriptsize{[ECCV16]}}~\citep{grounder}+LCNet \textcolor{lightgray}{\scriptsize{[IEEE17]}}~\citep{lcnet} & - & 4.17 & 3.28 & 1.05 & - & - & - & - \\
    MATN \textcolor{lightgray}{\scriptsize{[CVPR18]}}~\citep{matn}+LCNet \textcolor{lightgray}{\scriptsize{[IEEE17]}}~\citep{lcnet} & - & 4.41 & 3.53 & 1.12 & - & - & - & - \\
    GroundeR \textcolor{lightgray}{\scriptsize{[ECCV16]}}~\citep{grounder}+CPL \textcolor{lightgray}{\scriptsize{[CVPR22]}}~\citep{cpl} &  - & 5.23 & 4.18 & 1.25 & - & - & - & -  \\
    RAIR \textcolor{lightgray}{\scriptsize{[CVPR21]}}~\citep{rair}+CPL \textcolor{lightgray}{\scriptsize{[CVPR22]}}~\citep{cpl} & - & 6.88 & 4.87 & 1.36 & - & - & - & -  \\ \hline
    \textit{Weakly-Supervised (Single-stage pipelines)} \\ \hline
    WSSTG \textcolor{lightgray}{\scriptsize{[ACL19]}}~\citep{chen-etal-2019-weakly} & 
 - & 6.52 & 4.54 & 1.27 & - & - & - & - \\
    AWGU \textcolor{lightgray}{\scriptsize{[ACMMM20]}}~\citep{Chen2020ActivitydrivenWS} & - & 8.20 & 4.48 & 0.78 & - & - & - & - \\
    Vis-CTX \textcolor{lightgray}{\scriptsize{[CVPR19]}}~\citep{nafae} & - & 9.76 & 6.81 &  1.03  & - & - & - & - \\
    WINNER \textcolor{lightgray}{\scriptsize{[CVPR23]}}~\citep{Li_2023_CVPR} & - & 14.20 & 17.24 & \underline{6.12} & - & - & - & - \\ 
    VCMA \textcolor{lightgray}{\scriptsize{[ECCV24]}}~\citep{vcma} & - & \underline{14.64} & \underline{18.60} & 5.75 & - & - & - & -\\ \hline
     W-GDINO (Ours-Baseline) & \underline{18.0} & 9.04 & 11.56 & 4.57 & \underline{23.3} & \underline{9.85} & \underline{13.30} & \underline{5.63}\\
     STPro  & \textbf{30.6} & \textbf{17.56} & \textbf{26.98} & \textbf{12.93} & \textbf{39.0} & \textbf{19.99} & \textbf{31.70} & \textbf{14.55} \\

     & \improve{~+12.6} & \improve{~+2.92} &\improve{~+8.38} & \improve{~+6.81} & \improve{~+15.7} & \improve{~+10.14} & \improve{~+18.40} & \improve{~+8.92}  \\ 
     \specialrule{1.5pt}{0pt}{0pt}
		\end{tabular}}
  \caption{Comparison with existing state-of-the-art weakly and fully-supervised methods on HCSTVG-1 and HCSTVG-2 datasets. \textbf{Bold} denotes best and \underline{underline} denotes second best.}
	\label{tab:sota_weakly_hcstvg}
\end{table*}

\begin{table*}[h]
	\centering
	\renewcommand{\arraystretch}{1.06}
	\scalebox{0.8}{
		\begin{tabular}{r cccc cccc}
			\specialrule{1.5pt}{0pt}{0pt}
			\rowcolor{mygray} 
			\cellcolor{mygray} & \multicolumn{4}{c}{ \cellcolor{mygray} Declarative Sentences} & \multicolumn{4}{c}{ \cellcolor{mygray}Interrogative Sentences} \\ 
			\rowcolor{mygray} 
			\multirow{-2}{*}{\cellcolor{mygray} Methods}  & m\_tIoU & m\_vIoU & vIoU@0.3 &  vIoU@0.5  & m\_tIoU & m\_vIoU & vIoU@0.3 &  vIoU@0.5  \\
			\hline
			\hline
    \textit{Fully-Supervised} \\ \hline
    Ground-R \textcolor{lightgray}{\scriptsize{[ECCV16]}}~\citep{Rohrbach2015GroundingOT}&  - & 9.8 & 11.0 & 4.1 & - & 9.3 & 11.4 & 3.2 \\
    STPR \textcolor{lightgray}{\scriptsize{[CVPR17]}}~\citep{stpr} & 34.6 & 10.1 & 12.4 & 4.3 & 33.7 & 10.0 & 11.7 & 4.4\\
    WSSTG \textcolor{lightgray}{\scriptsize{[ACL19]}}~\citep{chen-etal-2019-weakly} & - & 11.4 & 14.6 & 5.9 & - & 10.7 & 13.9 & 5.3  \\
   STGRN \textcolor{lightgray}{\scriptsize{[CVPR20]}}~\citep{vidstg} & 48.5 & 19.8 & 25.8 & 14.6 & 46.9 & 18.3 & 21.1 & 12.8 \\
    STVGBert \textcolor{lightgray}{\scriptsize{[ICCV21]}}~\citep{stvgbert} & - & 24.0 & 30.9 & 18.4 & - & 22.5 & 26.0 & 16.0 \\
    TubeDETR \textcolor{lightgray}{\scriptsize{[CVPR22]}}~\citep{Yang2022TubeDETRSV} & 48.1 & 30.4 & 42.5 & 28.2 & 46.9 & 25.7 & 35.7 & 23.2 \\
    STCAT \textcolor{lightgray}{\scriptsize{[NeurIPS22]}}~\citep{Jin2022EmbracingCA} & 50.8 & 33.1 & 46.2 & 32.6 & 49.7 & 28.2 & 39.2 & 26.6  \\
    CSDVL \textcolor{lightgray}{\scriptsize{[CVPR23]}}~\citep{clb} & - & 33.7 & 47.2 & 32.8 & - & 28.5 & 39.9 & 26.2  \\
   CG-STVG \textcolor{lightgray}{\scriptsize{[CVPR24]}}~\citep{cg-stvg} & 51.4 & 34.0 & 47.7 & 33.1 & 49.9 & 29.0 & 40.5 & 27.5  \\
    VGDINO \textcolor{lightgray}{\scriptsize{[CVPR24]}}~\citep{video-gdino} & 52.0 & 34.7 & 48.1 & 34.0 & 50.8 & 29.9 & 41.0 & 27.6 \\ 
    \hline       
    \textit{Weakly-Supervised (Two-stage pipelines)} \\ \hline
    GroundeR\textcolor{lightgray}{\scriptsize{[ECCV16]}}~\citep{grounder}+LCNet  \textcolor{lightgray}{\scriptsize{[IEEE17]}}~\citep{lcnet} & - & 7.85 & 7.96 & 3.02 & - & 6.43 & 6.58 & 2.92 \\
    MATN\textcolor{lightgray}{\scriptsize{[CVPR18]}}~\citep{matn}+LCNet \textcolor{lightgray}{\scriptsize{[IEEE17]}}~\citep{lcnet} & - & 8.16 & 8.03 & 3.59 & - & 6.97 & 6.64 & 3.05 \\
    GroundeR\textcolor{lightgray}{\scriptsize{[ECCV16]}}~\citep{grounder}+CPL\textcolor{lightgray}{\scriptsize{[CVPR22]}}~\citep{cpl} & - & 8.28 & 8.35 & 3.68 & - & 7.16 & 7.28 & 3.23 \\
    RAIR \textcolor{lightgray}{\scriptsize{[CVPR21]}}~\citep{rair}+CPL \textcolor{lightgray}{\scriptsize{[CVPR22]}}~\citep{cpl} & - & 8.67 & 8.72 & 4.01 & - & 7.68 & 7.71 & 3.58 \\ \hline
    \textit{Weakly-Supervised (Single-stage pipelines)} \\ \hline
    WSSTG \textcolor{lightgray}{\scriptsize{[ACL19]}}~\citep{chen-etal-2019-weakly} & - & 8.85 & 8.52 & 3.87 & - & 7.12 & 6.87 & 2.96 \\
    AWGU \textcolor{lightgray}{\scriptsize{[ACMM20]}}~\citep{Chen2020ActivitydrivenWS} & - & 8.96 & 7.86 & 3.10 & - & 8.57 & 6.84 & 2.88 \\
    Vis-CTX \textcolor{lightgray}{\scriptsize{[CVPR19]}}~\citep{nafae} & - & 9.34 & 7.32 & 3.34 & - & 8.69 & 7.18 & 2.91   \\
    WINNER \textcolor{lightgray}{\scriptsize{[CVPR23]}}~\citep{Li_2023_CVPR} & - & 11.62 & 14.12 & 7.40 & - & 10.23 & 11.96 & 5.46  \\ 
    VCMA  \textcolor{lightgray}{\scriptsize{[ECCV24]}}~\citep{vcma} & - & \underline{14.45} & \underline{18.57} & \underline{8.76} & - & \textbf{13.25} & \textbf{16.74} & \underline{7.66} \\ \hline
    W-GDINO (Ours-Baseline)  & \underline{28.7} & 10.69 & 13.02 & 7.83 & \underline{29.1} & 9.87 & 12.16 & 6.71  \\
     STPro (Ours) &  \textbf{35.8} & \textbf{15.52} & \textbf{19.39} & \textbf{12.69}  & \textbf{34.6} & \underline{12.56} & \underline{14.95} & \textbf{9.29}\\
     & \improve{~+7.1} & \improve{~+1.07} & \improve{~+0.82} & \improve{~+3.93}  & \improve{~+5.5} & \improve{~-0.69} & \improve{~-1.79} & \improve{~+1.63}\\
    	\specialrule{1.5pt}{0pt}{0pt}
	\end{tabular}}
 \caption{Comparison with existing state-of-the-art weakly and fully-supervised methods on VidSTG dataset. \textbf{Bold} denotes best and \underline{underline} denotes second best.}
	\label{tab:sota_weakly_vidstg}

\end{table*}

\section{Comparison with Fully-Supervised Approaches}
\label{sec:stpro_vs_fully_sup}

We show a comparison of STPro with fully-supervised approaches (See Table \ref{tab:sota_weakly_hcstvg} and Table \ref{tab:sota_weakly_vidstg}). In comparison to recent approaches\cite{clb, cg-stvg, video-gdino}, our approach performs nearly at 50\% on both mean vIoU and tIoU on HCSTVG-1, HCSTVG-2 and VidSTG.

Though WSSTVG involves joint spatio-temporal inference, no existing work shows performance on the tIoU performance metric. We also include our tIoU scores for this task.   

\section{Joint-Inference Ablations}
\label{sec:inference_analysis}
STPro’s joint inference pipeline consists of two stages: First, TRM filters and trims tubelet proposals to a subset of temporall refined candidates; Then, SRM grounds the correct tubelet based on query-tubelet similarity from the selected candidates. In Section \ref{sec:trm_as_filter}, we present an ablation on the tubelet filtering criteria used in TRM. CG-SCL employs soft-label filtering (SLF) to obtain more relevant training tubelets for SRM, which can also be applied at inference time in tandem with TRM candidate selection. We explore the impact of this added filtering in Section~\ref{sec:slf_at_inference_time}. For VidSTG, we find that using TRM for both filtering and trimming reduces the overall referral performance. Hence, we use TRM only as a filter as discussed in Section ~\ref{sec:trm_for_vidstg}.

\subsection{TRM for Filtering \& Trimming}
\label{sec:trm_as_filter} 
\begin{figure}[H]
    \centering
    \includegraphics[width=0.9\linewidth]{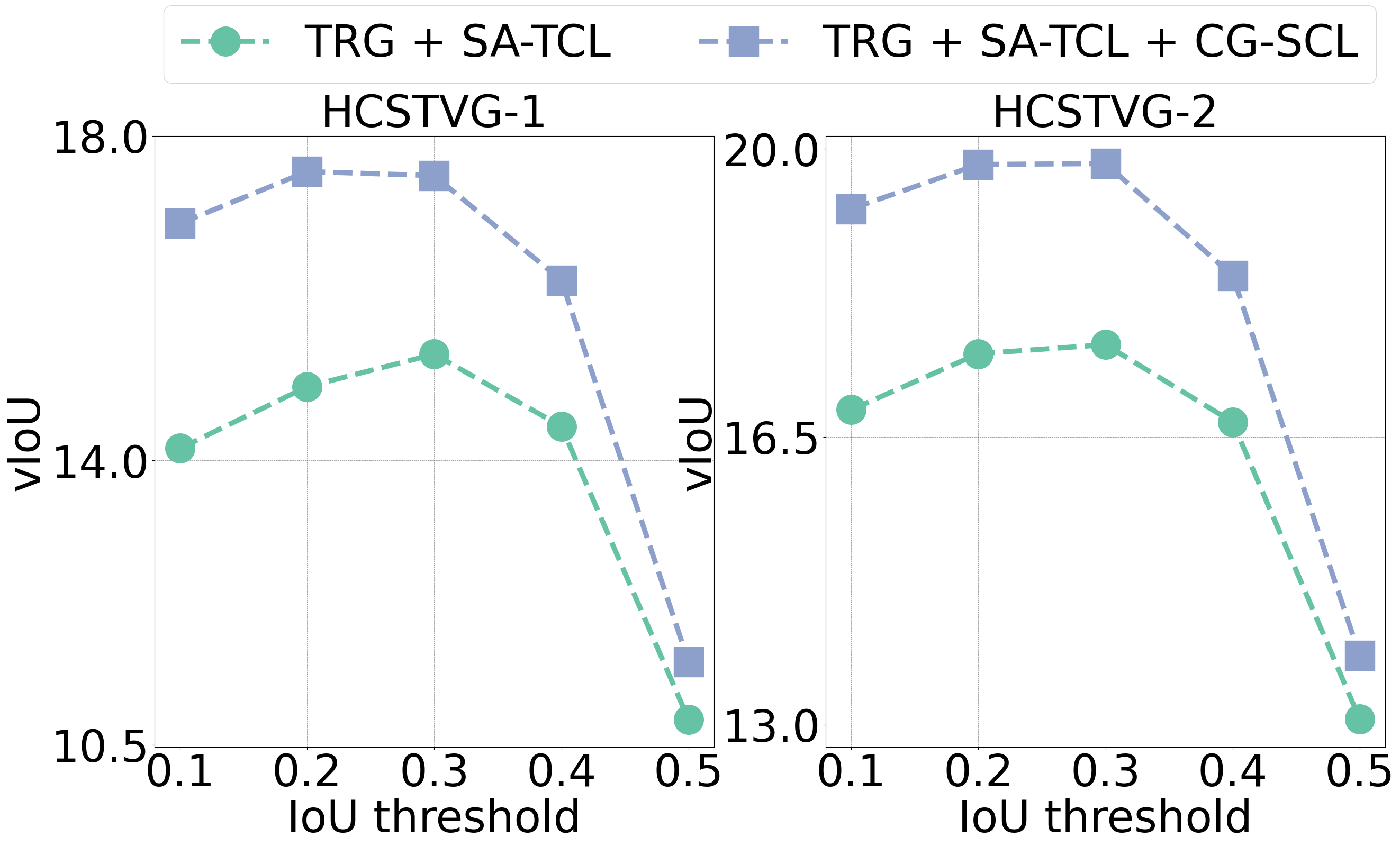}
    \caption{Joint spatio-temporal performance at different IoU thresholds ($T_{filt}$) for TRM on HCSTVG-1 \& HCSTVG-2.}
    \label{fig:iou_threshold}
\end{figure}
\begin{figure}[h]
    \centering
    \includegraphics[width=0.9\linewidth]{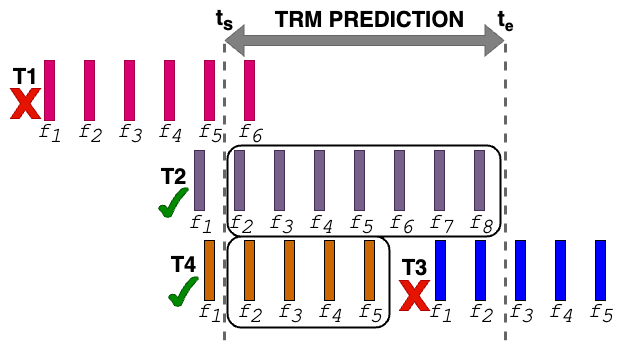}
    \caption{TRM for tubelet filtering \& trimming}
    \label{fig:tubelet_trimming}
\end{figure}
To select candidate tubelets from all proposals, TRM generates a temporal prediction $[t_s, t_d]$, where $t_s$ is the start time and $t_d$ is the end time. A tubelet is considered a valid candidate if it either fully contains or is fully contained within the TRM prediction. For tubelets that partially overlap with TRM’s prediction, we compute their temporal IoU and select those with an IoU above a threshold $T_{filt}$, ensuring only temporally relevant tubelets are retained, as shown in Figure~\ref{fig:tubelet_trimming}. The joint inference results at different $T_{filt}$ thresholds are shown in Figure~\ref{fig:iou_threshold}. Before being used in SRM, selected tubelets undergo nearest-neighbor interpolation to fill missing frames and are trimmed to the predicted temporal boundaries.

We find that $T_{filt} = 0.2$ works best for STPro, while $T_{filt} = 0.3$ is optimal for SA-TCL. As TRM’s temporal grounding capabilities improve, we expect the threshold $T_{filt}$ to increase for optimal performance.

\subsection{SLF at Inference Time}
\label{sec:slf_at_inference_time}
\begin{table}[H]
    \centering
    \renewcommand{\arraystretch}{1.1}
    \scalebox{0.77}{
        \begin{tabular}{c | c | ccc  ccc}
            \specialrule{1.5pt}{0pt}{0pt}
            \rowcolor{mygray} 
   \rowcolor{mygray} Dataset & SLF &  m\_tIoU & m\_vIoU & vIoU@0.1&vIoU@0.3&vIoU@0.5 \\ \hline\hline
    HCSTVG-1 & & \textbf{30.81} & 17.37 & 41.81 & 26.55 & 12.67 \\
    HCSTVG-1 & \checkmark & 30.56 &\textbf{17.56} & \textbf{41.90} & \textbf{26.98} & \textbf{12.93} \\ \hline
    HCSTVG-2 & & \textbf{38.99} & 19.53 & 47.45 & 30.85 & 13.8\\
    HCSTVG-2 & \checkmark & 38.46 & \textbf{19.82} & \textbf{47.95} & \textbf{31.35} & \textbf{14.25} \\ 
    \specialrule{1.5pt}{0pt}{0pt}
			\end{tabular}}
    \caption{STPro performance with and without soft-label-filtering at inference time on HCSTVG-1 and HCSTVG-2.}
 	\label{tab:slf_at_inference_time}
\end{table}
We apply SLF, as described in Section~\ref{sec:slf_in_practice}, as an additional filter to the candidate tubelets selected by TRM. The results with and without test-time SLF are shown in Table~\ref{tab:slf_at_inference_time}. We find that SLF at test time provides minimal improvement, and our models outperform previous baselines even without SLF during inference. This suggests that TRM alone effectively filters the candidate tubelets, enabling better referral grounding in SRM. Dataset statistics in Figure~\ref{fig:dataset_stats_r1} further support this, showing that few samples have a large number of temporally overlapping tubelets (stage-1 and stage-2), reducing the need for additional filtering.

\subsection{TRM as a Filter for VidSTG}
\label{sec:trm_for_vidstg}
\begin{table}[H]
	\centering
\renewcommand{\arraystretch}{1.1}
\scalebox{0.8}{
\begin{tabular}{ccc | ccccc}
    \rowcolor{mygray} 
    \specialrule{1.5pt}{0pt}{0pt}
    \rowcolor{mygray} 
    SRM & Filter & Trim & m\_tIoU & m\_vIoU & vIoU@0.1 &  vIoU@0.3 &  vIoU@0.5  \\ 
    \hline\hline
          &&& 28.79 & 10.69 & 25.03& 13.02 & 7.83\\
        \checkmark & & & 34.17 & 14.93 & 32.49 & 18.44 & 12.26 \\
        \checkmark & \checkmark  & \checkmark & 24.97 & 11.1 & 24.97 & 17.29 & 4.75 \\
        \checkmark & \checkmark & & \textbf{35.77} & \textbf{15.52} & \textbf{33.41} & \textbf{19.39} & \textbf{12.69} \\
    \specialrule{1.5pt}{0pt}{0pt}
    \rowcolor{mygray} 
    SRM & Filter & Trim  & m\_tIoU & m\_vIoU & vIoU@0.1 & vIoU@0.3 &  vIoU@0.5  \\ 
    \hline\hline 
    &&& 29.10 & 9.87 & 23.19 &  12.16 & 6.71 \\
    \checkmark & & & 33.64 & 12.29 & 28.3 & 14.56 & 9.05 \\
    \checkmark & \checkmark  & \checkmark & 24.39 & 8.76 & 24.86 & 12.4 & 2.85\\
    \checkmark & \checkmark & & 
    \textbf{34.64} & \textbf{12.56} & \textbf{29.12} & \textbf{14.95} & \textbf{9.29} \\

 \specialrule{1.5pt}{0pt}{0pt}
\end{tabular}}
  \caption{Analysis on the use of TRM as filter and trimmer versus only as a filter for joint spatiotemporal inference on VidSTG declarative (upper), VidSTG interrogative (lower).}
    \label{tab:vidstg_trim_vs_filter}
\end{table}
Using TRM as both a filter and trimmer leads to a performance decline when combined with SRM for both VidSTG declarative and VidSTG interrogative tasks. Analysis shows that VidSTG tubelet proposals, generated through detection and tracking, as well as their ground-truth temporal boundaries, are heavily skewed toward durations under 2 seconds. In contrast, TRM often predicts significantly larger temporal spans. Following \cite{cnm}, instead of imposing an upper limit on temporal width predictions, we avoid further trimming already short tubelets. We apply nearest-neighbor interpolation but restrict TRM to act only as a filter. The performance decline from trimming and the improvement from using TRM purely as a filter are shown in Table~\ref{tab:vidstg_trim_vs_filter}.

\section{Soft-Label Filtering (SLF) in Practice}
\label{sec:slf_in_practice}
\begin{figure*}[h]
    \centering
    \includegraphics[width=0.9\linewidth]{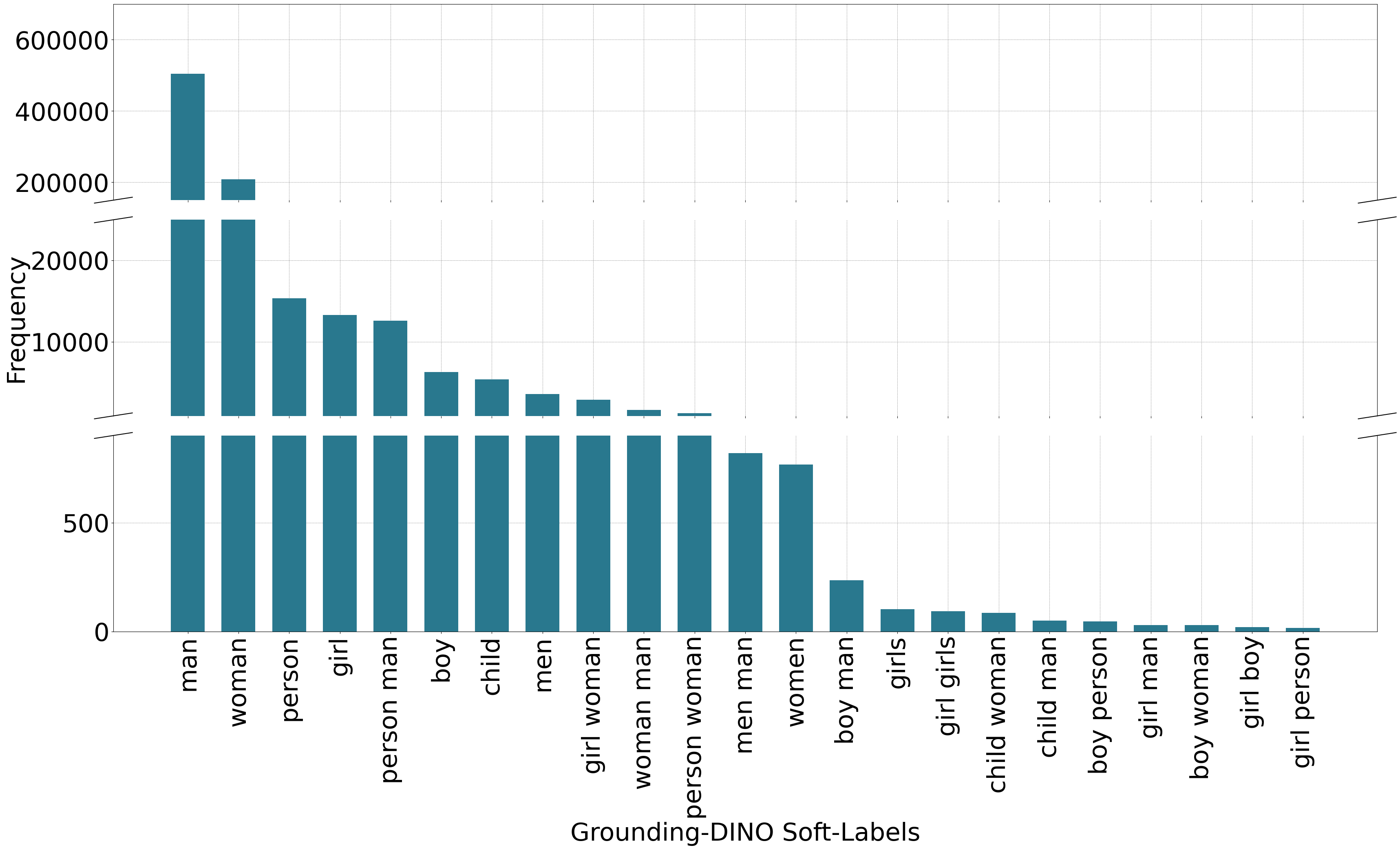} 
    \caption{ G-DINO soft-label distribution HCSTVG-1.}
    \label{fig:gdino_soft_label_dist_hcstvg1}
\end{figure*}

\begin{figure*}[h]
    \centering
    \includegraphics[width=0.9\linewidth]{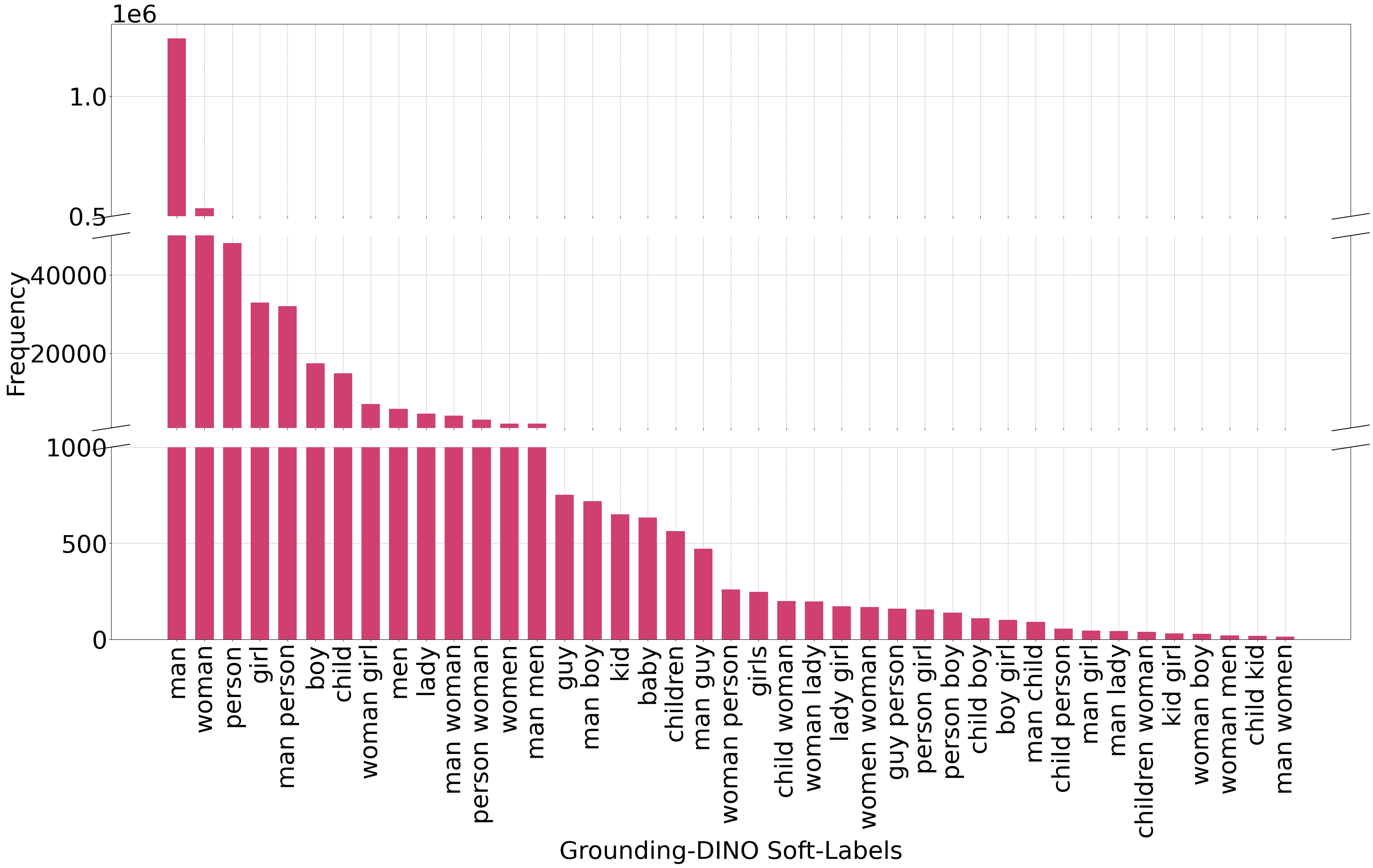} 
    \caption{ G-DINO soft-label distribution HCSTVG-2.}
    \label{fig:gdino_soft_label_dist_hcstvg2}
\end{figure*}

To enhance TRM's ability to distinguish referral subjects based on attributes and subject types, we employ Soft Label Filtering (SLF) in CG-SCL. Figures \ref{fig:gdino_soft_label_dist_hcstvg1} and \ref{fig:gdino_soft_label_dist_hcstvg2} illustrate the distribution of soft labels obtained from Grounding-DINO detections across all tubelets in the training set for HCSTVG-1 and HCSTVG-2, respectively. 

In Figure~\ref{fig:gdino_soft_label_dist_hcstvg2}, we observe two distinct types of labels: straightforward labels such as \emph{"man"} and \emph{"woman"}, which represent unambiguous categories, and compound labels such as \emph{"person woman"} and \emph{"man woman"}, which occur frequently. To enable proper tubelet selection, we use \emph{gender} as a filtering criterion. 

The filtering rules are as follows:
\begin{itemize}
    \item For compound labels such as \emph{"woman person"}, the gender is assigned based on the specific component (e.g., \emph{"woman"}).
    \item For ambiguous labels such as \emph{"child"}, or compound labels like \emph{"man woman"} and \emph{"boy girl"}, we assign a new \emph{neutral gender} to unify all such uncertain or generic labels.
    \item For each tubelet, the dominant gender across all individual detections is used as its label.
\end{itemize}

Additionally, the majority of referred subjects in captions belong to categories such as \emph{man}, \emph{woman}, \emph{person}, \emph{child}, \emph{boy}, \emph{girl}, \emph{kid}, and \emph{lady}. While these subjects have identifiable genders, to account for other specific subjects like \emph{singer} or \emph{police officer}, we consider them as being gender neutral. For filtration, we then apply the following rules:
\begin{itemize}
    \item If the referred subject (in caption) is neutral: Include all tubelets for training.
    \item If the referred subject (in caption) is specific: Include all tubelets with matching gender as well as those with \emph{neutral gender}.
\end{itemize}

\subsubsection*{Extending the Approach}
This method is flexible and can be adapted to other datasets:
\begin{itemize}
    \item Each category can be treated independently.
    \item If a detection combines parent and child categories (e.g., \emph{"woman person"}), it can be assigned the child label (e.g., \emph{"woman"}).
    \item For combinations of two or more specific categories (e.g., \emph{"man woman"}), a \emph{neutral label} can be assigned, ensuring these tubelets are always included for training.
\end{itemize}

\begin{figure*}[h]
    \centering
    \includegraphics[width=\linewidth]{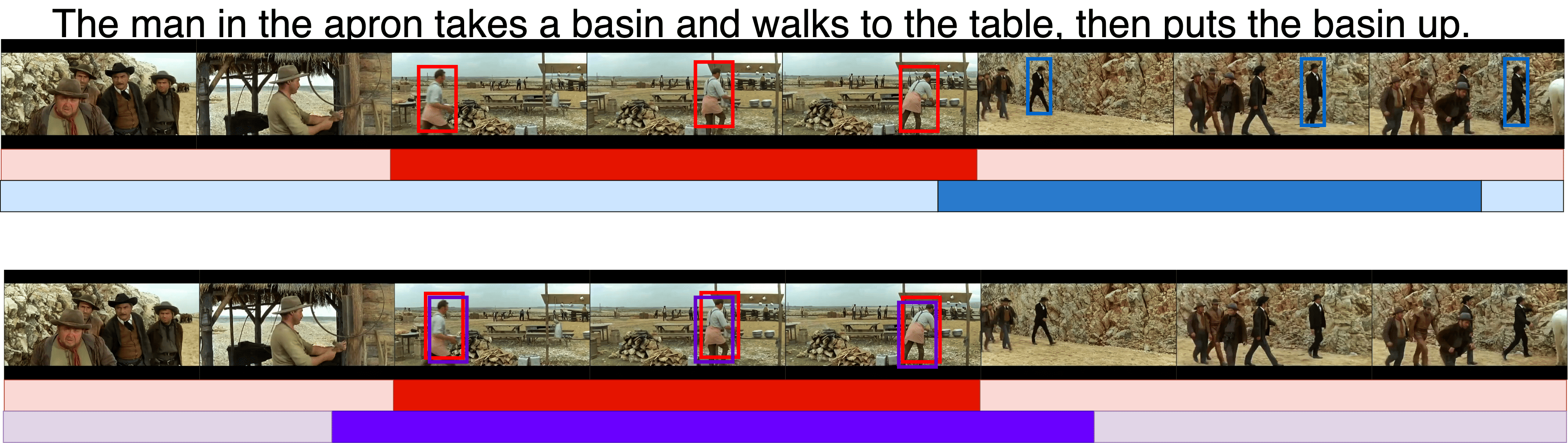} 
    \caption{\textbf{Qualitative Impact of SA-TCL:} Red indicates the ground-truth, Blue represents the prediction from TRG, and Violet corresponds to STPro (SA-TCL + CG-SCL applied over TRG). Darker shades denote predictions. The bars below the frames depict the temporal predictions from TRM, while the bounding boxes represent the tubelets grounded by SRM. In the top example, TRM incorrectly localizes the query temporally, leading to a failure in SRM’s spatial grounding. In contrast, the bottom example demonstrates how STPro significantly refines TRM’s temporal predictions, enabling SRM to correctly ground the referred subject.}
    \label{fig:tcl_qual}
\end{figure*}

\begin{figure*}[h]
    \centering
    \includegraphics[width=\linewidth]{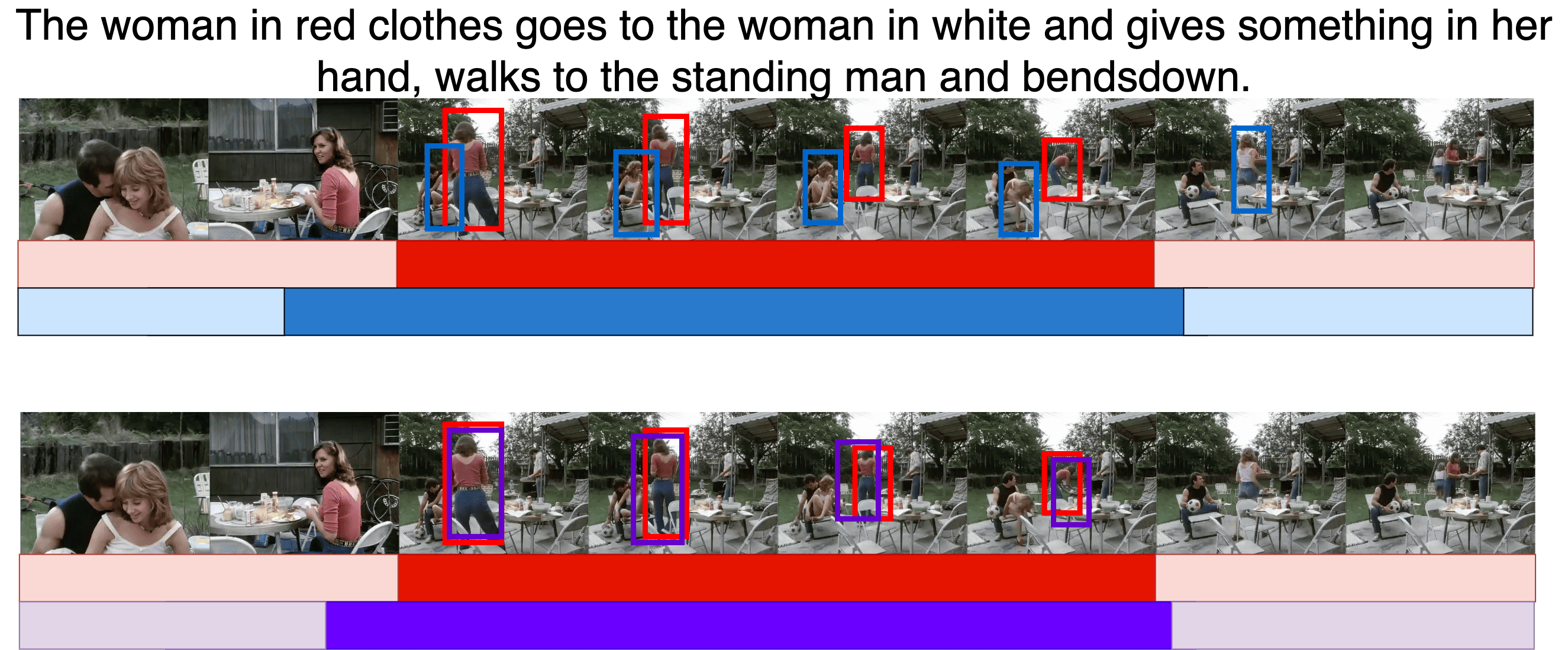} 
    \caption{\textbf{Qualitative Impact of CG-SCL:} Red indicates the ground truth, Blue represents the prediction from TRG, and Violet corresponds to STPro (SA-TCL + CG-SCL applied over TRG). Darker shades denote predictions. The bars below the frames depict the temporal predictions from TRM, while the bounding boxes represent the tubelets grounded by SRM. In the top example, SRM incorrectly localizes the referred subject (\textit{woman in red clothes}) though TRM's prediction is near optimal. In contrast, the bottom example demonstrates how STPro significantly refines SRM to correctly ground the referred subject.}
    \label{fig:scl_qual}
\end{figure*}

\section{Data Pre-processing Prompts}
\label{sec:pre_proc_prompts}

\newtcolorbox{promptbox}{
    colback=white,
    colframe=black,
    sharp corners,
    boxrule=0.5mm,
    fonttitle=\bfseries,
    enlarge left by=0mm,
    enlarge right by=0mm,
    breakable
}

\lstdefinelanguage{json}{
    basicstyle=\ttfamily\small,
    numbers=none,
    stepnumber=1,
    numbersep=8pt,
    showstringspaces=false,
    breaklines=true,
    frame=none,
    literate=
     *{0}{{{\color{blue}0}}}{1}
      {1}{{{\color{blue}1}}}{1}
      {2}{{{\color{blue}2}}}{1}
      {3}{{{\color{blue}3}}}{1}
      {4}{{{\color{blue}4}}}{1}
      {5}{{{\color{blue}5}}}{1}
      {6}{{{\color{blue}6}}}{1}
      {7}{{{\color{blue}7}}}{1}
      {8}{{{\color{blue}8}}}{1}
      {9}{{{\color{blue}9}}}{1}
}

\subsection{Sub-Action Extraction}
\label{sec:sub_action_extraction}
The prompt used with GPT 3.5-Turbo for sub-action extraction for each caption in the train and test set.

\begin{promptbox}
\textbf{System:} You are a language expert. Your task is to break up a sentence into multiple sub-sentences. Any given sentence may contain multiple verbs/actions. Each sub-sentence will contain some subset of the actions in the original sentence. When splitting the sentence into sub-sentences, you must ensure that the actions are only grouped in the order in which they appear in the original sentence. You cannot skip over any actions in the original sentence; they should always be contiguous when splitting sentences. Your goal is to create all the different combinations of actions possible while maintaining their ordering. If a sentence has 10 actions, then we can create groups of up to 9 continuous actions. You must create all such groups. The keys within the JSON object are the number of actions used to form the group of sub-sentences, and the list within it is the list of sentences containing exactly those many actions. If some pronouns are ambiguous, qualify them with the object or subject they actually refer to. Each sentence must make complete sense on its own.

\textbf{\#\# Example 1} \\
\textbf{Original sentence:} The man in the jacket walks to the woman in red and stops, takes a golf club, gives it to the woman in red, and pushes her. \\
\textbf{Answer:}
\begin{lstlisting}[language=json]
{
    "1": [
        "The man in the jacket walks to the woman in red",
        "The man in the jacket stops",
        "The man in the jacket takes a golf club",
        "The man in the jacket gives the golf club to the woman in red",
        "The man in the jacket pushes her"
    ],
    "2": [
        "The man in the jacket walks to the woman in red and stops",
        "The man in the jacket stops, takes a golf club",
        "The man in the jacket takes a golf club, gives it to the woman in red",
        "The man in the jacket gives the golf club to the woman in red, and pushes her"
    ],
    "3": [
        "The man in the jacket walks to the woman in red and stops, takes a golf club",
        "The man in the jacket stops, takes a golf club, gives it to the woman in red",
        "The man in the jacket takes a golf club, gives it to the woman in red, and pushes her"
    ],
    "4": [
        "The man in the jacket walks to the woman in red and stops, takes a golf club, gives it to the woman in red",
        "The man in the jacket stops, takes a golf club, gives it to the woman in red, and pushes her"
    ]
}
\end{lstlisting}

\textbf{\#\# Example 2} \\
\textbf{Original sentence:} The bald man leaves the room, pulls the door, walks towards the man in the white suit, and then turns to face the white-suit man. \\
\textbf{Answer:}
\begin{center}
    \huge \textbf{...} \\
    \vspace{0.1cm}
\end{center} 
\textbf{Original sentence:} \textit{<train/test caption>} \par 
\textbf{Answer:} \\
\end{promptbox}
\textbf{Failure case:} Though the actions and their relative orders which comprise sub-action phrases are largely correct, at times the number of combinations of extracted sub-action phrases is incorrect. For example : 
\begin{quotation}
\centering
\begin{minipage}{1.0\linewidth}
    {\texttt{$Q$: The man sitting on the left in the first row in brown clothes \underline{stands} up, \underline{waves} his hands and \underline{talks}, \underline{turns} around, \underline{looks} down, then \underline{sits} down again.}} \\
    \newline
    {\texttt{$Answer$:}}
    \begin{verbatim}
    {
        "1": [...],
        "2": [...],
        "3": [...],
        "4": [...]
    }
    \end{verbatim} \\
\end{minipage}
\end{quotation}
In this case, we expect five keys to be present in the answer JSON since we can make two combinations of five actions in order. However, we get only four keys in the response, effectively losing out on some training examples in our devised SA-TCL.

\subsection{Descriptor (POS) Extraction}
\label{sec:pos_extraction}
The prompt used with GPT 3.5-Turbo for POS extraction for each caption in the train and test set.

\begin{promptbox}
\textbf{System:} Extract the quantifier phrase describing the main person.

\textbf{\#\# Example 1} \\
\textbf{Original sentence:} The man in brown clothes pours the contents of the bag into his hand, and then takes out a piece of paper from the bag and opens it. \\
\textbf{Answer:} The main person in this sentence is \"The man in brown clothes.\" \\

\textbf{\#\# Example 2} \\
\textbf{Original sentence:} The girl gets up, is caught by the opposite man, and pushes her to sit down. \\
\textbf{Answer:} The main person in this sentence is \"The girl.\" \\

\textbf{\#\# Example 3} \\
\textbf{Original sentence:} The man in the hat reaches out and points to the front, then puts his hand down and turns his head. \\
\textbf{Answer:} The main person in this sentence is \"The man in the hat.\" \\
\begin{center}
    \vfill
    \huge \textbf{...} \\
    \vspace{0.1cm}
\end{center} 

\textbf{Original sentence:} \textit{<train/test caption>} \\
\textbf{Answer:}
\end{promptbox}

\textbf{Failure case:} The extracted referral subject is nearly always correct, however in some cases, the referral subject itself does not contain any distinguishable attributes. Instead the series of actions performed serves as the distinguishing factor, which our extraction scheme fails to capture. For example : 
\begin{quotation}
\centering 
\begin{minipage}{1.0\linewidth}
    {\texttt{$Q$: <prompt> The woman who points her hand at the other woman}} \\
    \newline
    {\texttt{$Answer$: The main person in this sentence is The woman}}
\end{minipage}
\end{quotation}

In this case, the identifying characteristic is the action itself. This would lead to the generation of a noisy training sample where-in any related tubelet might be selected but the incorrect one for the particular sample.

\section{Qualitative Analysis (STPro)}
\label{sec:qualitative_analysis}

STPro improves model performance for weakly supervised temporal video grounding (WSTVG) in two key ways:

\textbf{Improved Temporal Prediction via SA-TCL:} As shown in Figure~\ref{fig:tcl_qual}, the original predictions from the temporal reasoning module (TRM) fail to include the optimal tubelet of the referred subject. Consequently, the candidate proposals obtained from TRM do not contain the referred subject, leading to an incorrect spatial prediction by the spatial reasoning module (SRM). By applying SA-TCL, the temporal predictions of TRM are refined, ensuring that the correct referral subject is included among the candidate proposals. In this case, the revised temporal boundary contains only one tubelet, corresponding to the referred subject, enabling accurate spatial grounding.

\textbf{Enhanced Referral Grounding via CG-SCL:} In scenarios where the temporal predictions of TRM and SA-TCL are nearly identical and near-optimal (Figure~\ref{fig:scl_qual}), SRM may still misidentify the referred subject due to insufficient grounding of distinguishing features, such as \textit{red}. For instance, the original SRM prediction incorrectly grounds another subject in the spatio-temporal region. CG-SCL directly enhances SRM’s referral grounding capabilities, enabling STPro to correctly ground the referred subject based on spatial and temporal cues.
By combining temporal refinement through SA-TCL and spatial enhancement via CG-SCL, STPro achieves significant improvements in WSTVG tasks.

\end{document}